\documentclass{article}

\usepackage{microtype}
\usepackage{graphicx}
\usepackage{subfigure}
\usepackage{booktabs} 
\usepackage{multirow}
\usepackage{colortbl}
\usepackage{xcolor}

\usepackage{hyperref}

\usepackage[accepted]{icml2024}
\usepackage{changdefs}

\newcommand{\appxref}[1]{Appendix~\ref{appx:#1}}
\newcommand{\data}{\textnormal{label}}
\newcommand{\SC}{\textnormal{self-con}}
\newcommand{\ext}{\textnormal{ext}}
\newcommand{\XC}{\textnormal{XC}}

\newcommand{\ee}{\textnormal{ee}}

\newcommand{\clHh}{\hat{\clH}}
\newcommand{\bfHh}{\hat{\bfH}}
\newcommand{\bfDt}{\tilde{\bfD}}
\newcommand{\scloss}{{self-con}}

\makeatletter
\newcommand{\subalign}[1]{%
  \vcenter{%
    \Let@ \restore@math@cr \default@tag
    \baselineskip\fontdimen10 \scriptfont\tw@
    \advance\baselineskip\fontdimen12 \scriptfont\tw@
    \lineskip\thr@@\fontdimen8 \scriptfont\thr@@
    \lineskiplimit\lineskip
    \ialign{\hfil$\m@th\scriptstyle##$&$\m@th\scriptstyle{}##$\hfil\crcr
      #1\crcr
    }%
  }%
}
\makeatother

\usepackage{amsmath}
\usepackage{amssymb}
\usepackage{mathtools}
\usepackage{amsthm}

\usepackage[textsize=tiny]{todonotes}

\icmltitlerunning{Self-Consistency Training for Density-Functional-Theory Hamiltonian Prediction}

\begin{document}

\twocolumn[
\icmltitle{\hspace{-1pt}Self-Consistency Training for Density-Functional-Theory Hamiltonian Prediction\hspace{-1pt}}




\begin{icmlauthorlist}
\icmlauthor{He Zhang}{xjtu,ai4s,intn}
\icmlauthor{Chang Liu}{ai4s}
\icmlauthor{Zun Wang}{ai4s}
\icmlauthor{Xinran Wei}{ai4s}
\icmlauthor{Siyuan Liu}{ai4s,intn}
\icmlauthor{Nanning Zheng}{xjtu}
\icmlauthor{Bin Shao}{ai4s}
\icmlauthor{Tie-Yan Liu}{ai4s}
\end{icmlauthorlist}

\icmlaffiliation{xjtu}{National Key Laboratory of Human-Machine Hybrid Augmented Intelligence, National Engineering Research Center for Visual Information and Applications, and Institute of Artificial Intelligence and Robotics, Xi'an Jiaotong University}
\icmlaffiliation{ai4s}{Microsoft Research AI for Science}
\icmlaffiliation{intn}{These authors did this work during an internship at Microsoft Research AI for Science}

\icmlcorrespondingauthor{Chang Liu}{changliu@microsoft.com}
\icmlcorrespondingauthor{Nanning Zheng}{nnzheng@mail.xjtu.edu.cn}

\icmlkeywords{AI for science, molecular property prediction, electronic structure, physics-informed training}

\vskip 0.3in
]



\printAffiliationsAndNotice{}

\begin{abstract}
  Predicting the mean-field Hamiltonian matrix in density functional theory is %
  a %
  fundamental formulation to leverage machine learning for solving molecular science problems.
  Yet, its applicability is limited by insufficient labeled data for training. %
  In this work, we highlight that Hamiltonian prediction possesses a self-consistency principle, based on which we propose self-consistency training, an exact training method that does not require labeled data.
  It distinguishes the task from predicting other molecular properties by the following benefits:
  \itemone~it enables the model to be trained on a large amount of unlabeled data, %
  hence addresses the data scarcity challenge and enhances generalization;
  \itemtwo~it is more efficient than running DFT to generate labels for supervised training, since it amortizes DFT calculation over a set of queries. %
  We empirically demonstrate the better generalization in data-scarce and out-of-distribution scenarios, and the better efficiency over DFT labeling.
  These benefits push forward the applicability of Hamiltonian prediction to an ever-larger scale.
\end{abstract}

\section{Introduction}
\label{sec:intro}

\begin{figure}
    \centering
    \includegraphics[width=0.9\columnwidth]{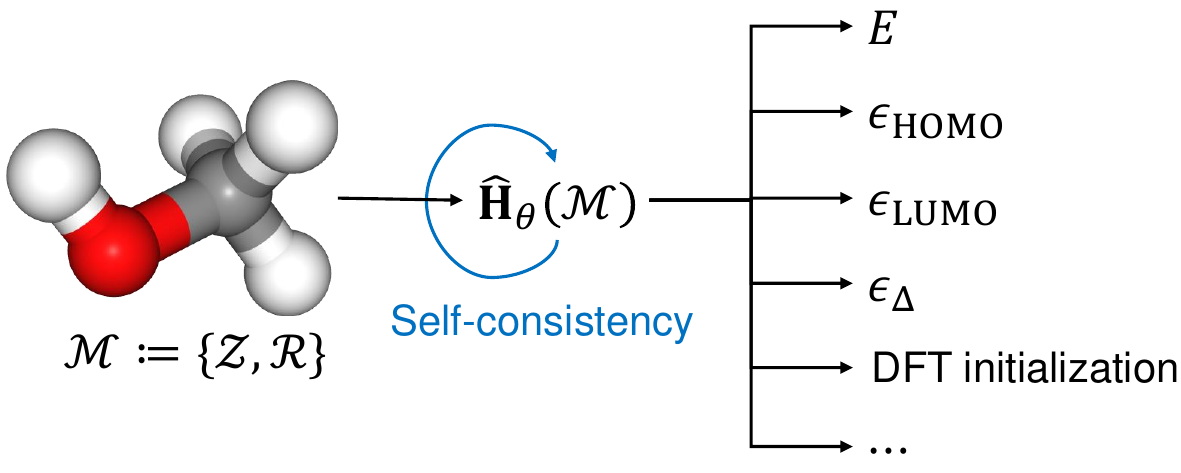}
    \vspace{-0.1in}
    \caption{Hamiltonian prediction is the task to use a machine learning model to predict the mean-field Hamiltonian matrix $\bfHh_\theta(\clM)$ in density functional theory from a given molecular structure $\clM := \{\clZ, \clR\}$ specified by the atomic types $\clZ$ and coordinates $\clR$ of atoms. It can derive various molecular properties, \eg, the total energy $E$, the HOMO and LUMO energies $\epsilon_\mathrm{HOMO}, \epsilon_\mathrm{LUMO}$ and their gap $\epsilon_\Delta$ for the given molecule, and can also serve as an accurate DFT initialization. We highlight in this work that the task has a self-consistency principle (the blue loop arrow), which allows training the model without labeled data.} %
    \label{fig:ham-pred}
    \vspace{-0.2in}
\end{figure}

Calculating properties of molecules is the foundation for a wide range of industry needs including drug design, protein engineering, and material discovery.
The key to these properties is the %
electronic structure in the molecule, for which various computational methods are proposed.
Density functional theory (DFT)~\citep{hohenberg1964inhomogeneous,kohn1965self,perdew1996generalized,teale2022dft} is perhaps the most prevailing choice due to its balanced accuracy and efficiency,
but still hard to meet the demand in industry.
Encouraged by the impressive advancement in machine learning, researchers have trained machine learning models on datasets with property labels to directly predict properties of queried molecules~\citep{ramakrishnan2014quantum,chmiela2019sgdml,chanussot2021open}.
For each property, a separate model (at least a separate output module) needs to be trained.
A more fundamental formulation is to predict the Hamiltonian matrix~\citep{schutt2019unifying}, or more precisely, the effective one-electron mean-field Hamiltonian matrix, \ie, the Fock matrix in a DFT calculation after convergence.
The Hamiltonian matrix can directly provide all the properties that a DFT calculation can (\figref{ham-pred}), waiving the need to specify the target property or train multiple models.
Moreover, Hamiltonian prediction can also accelerate running DFT by providing an accurate initialization. %

Noticeable progress has been made for Hamiltonian prediction. \citet{hegde2017machine} pioneered the direction using kernel ridge regression to predict semi-empirical Hamiltonian for one-dimensional systems. \citet{schutt2019unifying} then proposed a neural network model called SchNorb to predict Hamiltonian for small molecules, which is further enhanced for prediction efficiency by \citet{gastegger2020deep}. \citet{shmilovich2022orbital} proposed to employ atomic orbital features for Hamiltonian prediction.
Noting that the Hamiltonian matrix is composed of tensors in various orders which are equivariant to coordinate rotation in respective ways, subsequent works proposed neural network model architectures that guarantee the equivariance. %
Some works~\citep{unke2021se,yu2023efficient,gong2023general,yin2024harmonizing} include high-order tensorial features %
into model input, which are processed in an equivariant way typically with tensor products.
\citet{li2022deep} used local frames to anchor coordinate systems with the molecule so that the prediction target is invariant.
\citet{zhang2022equivariant,nigam2022equivariant} implemented the prediction by constructing equivariant kernels.
There are works that exploited data other than the Hamiltonian directly, \eg, using orbital energies~\citep{wang2021machine,gu2022neural,zhong2023transferable} to supervise the prediction of Hamiltonian.
While these prior efforts have introduced powerful architectures showing encouraging outcomes, they all rely on datasets providing Hamiltonian or orbital energy labels. Since such datasets are scarce, %
the applicability of Hamiltonian prediction is restricted to molecules with no more than 31 atoms~\cite{yu2023qh9}.\footnote{
There are a few works~\citep{li2022deep,gong2023general} that have demonstrated applicability to large-scale material systems. We note that this is achieved by leveraging the periodicity and locality in material systems, %
which do not hold perfectly in molecular systems.
} %

In this work, we highlight a uniqueness of Hamiltonian prediction: it has a self-consistency principle (indicated by the blue loop arrow in \figref{ham-pred}), by leveraging which we design %
a training method that guides the model \emph{without labeled data}.
The self-consistency originates from the %
basic equation of DFT (\eqnref{ks-eq-mat}) that the Hamiltonian needs to satisfy. Conventional DFT solves the equation using a fixed-point iteration process called self-consistent field (SCF) iteration. In contrast, the proposed self-consistency training solves the equation by directly minimizing the %
residue of the equation incurred by the model-predicted Hamiltonian (\figref{sc-loss}). %
As the equation fully determines the prediction target, no Hamiltonian label is required, and the loss function is minimized only if the equation is satisfied and the prediction is exact.
Self-consistency training compensates data scarcity with physical laws, and differentiates Hamiltonian prediction from other machine learning formulations (\eg, energy prediction), in that it enables continued self-improvement without additional labeled data.

We exploit the merit of self-consistency training in two specific points.
\itemone Self-consistency training leverages unlabeled data, which allows substantial improvement of the \emph{generalizability} of the Hamiltonian prediction model. %
We demonstrate that the predicted Hamiltonian as well as derived molecular properties are indeed improved by a significant margin, when labeled data is limited (data-scarce scenario) and when the model is evaluated on molecules larger than those used in training (out-of-distribution scenario).

\itemtwo Self-consistency training on unlabeled data is more efficient %
than generating labels using DFT on those data for supervised learning,
as we find that self-consistency training can be seen as an \emph{amortization} of DFT calculation over a set of molecules.
DFT requires multiple SCF iterations on each molecule before providing supervision, while self-consistency training only requires effectively one SCF iteration to return a training signal,
hence can provide information on more molecules given the same amount of computation.
The better efficiency for Hamiltonian prediction training is empirically verified in both data-scarce and out-of-distribution scenarios.
More attractively, regarding physical quantities derived from the predicted Hamiltonian, self-consistency training even outperforms supervised training using full additional labels given sufficient computational budget, indicating that it is more relevant to molecular properties and real applications. %
We also verified the direct acceleration by self-consistency training to solve a bunch of molecules upon the conventional DFT calculation.

Finally, we demonstrate that with the above %
two unique benefits of self-consistency training, the applicability of Hamiltonian prediction can overcome the data limit, and is extended to molecules much larger (56 atoms) than previously reported, showing increased practical relevance.
It also derives orders better molecular property results on these large molecules than end-to-end property prediction models, which are always limited by the availability of labeled data.

\section{Self-Consistency Training} \label{sec:mth}

\subsection{Preliminaries}
\label{sec:prelim}

\begin{figure*}[ht]
    \centering
    \includegraphics[width=0.8\linewidth]{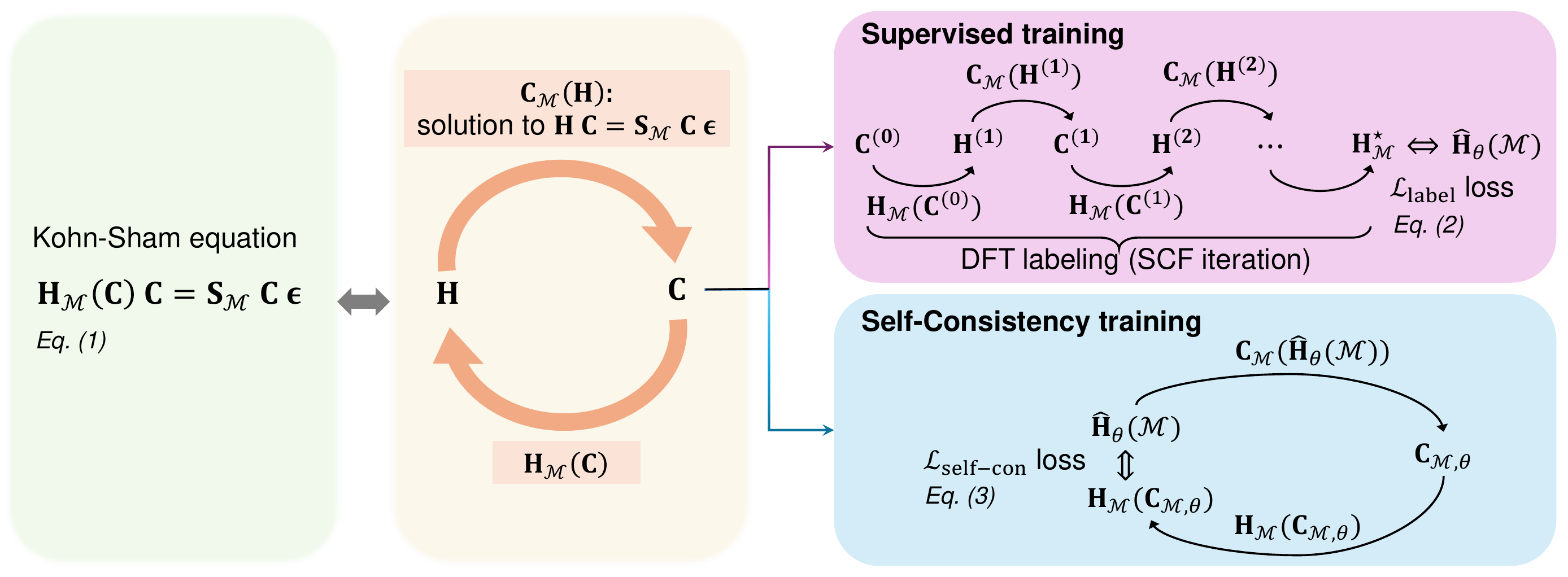}
    \caption{%
    Illustration of the proposed self-consistency training with comparison to the conventional DFT calculation and supervised training.
    (\textbf{Left}) The central task of a DFT calculation is to solve the Kohn-Sham equation (\eqnref{ks-eq-mat}) for the given molecular structure $\clM$. (\textbf{Middle}) The equation is equivalent to the condition that the eigenvectors $\bfC$ of $\bfH$ recover $\bfH$ via a known function $\bfH_\clM(\bfC)$.
    (\textbf{Top-Right})~To solve the equation, conventional DFT uses a fixed-point iteration (SCF iteration), which, upon convergence, gives the label $\bfH^\star_\clM$ for supervised training (\eqnref{data-loss}) of a Hamiltonian prediction model $\bfHh_\theta(\clM)$.
    (\textbf{Bottom-Right}) In contrast, self-consistency training (\eqnref{sc-loss}) directly minimizes the mismatch between the predicted Hamiltonian $\bfHh_\theta(\clM)$ and the matrix $\bfH_\clM(\bfC_{\clM,\theta})$ reconstructed from its eigenvectors.
    }
    \label{fig:sc-loss}
    \vspace{-0.05in}
\end{figure*}

We first provide a schematic description of the calculation mechanism of DFT and conventional supervised learning for Hamiltonian prediction before delving into self-consistency training. \appxref{dft} provides more details.

For a given molecular structure $\clM := \{\clZ, \clR\}$, where $\clZ := \{Z^{(a)}\}_{a=1}^A$ and $\clR := \{\bfR^{(a)}\}_{a=1}^A$ specify the atomic numbers (types) and coordinates of the $A$ nuclei in the molecule, DFT solves the ground state of the $N$ electrons in the molecule by minimizing electronic energy under a reduced representation of electronic state, which is $N$ one-electron wavefunctions %
$\{\phi_i(\bfrr)\}_{i=1}^N$, called orbitals. Here, $\bfrr \in \bbR^3$ represents the Cartesian coordinates of an electron.
For practical calculation, a basis set of functions on $\bbR^3$ is used to expand the orbitals. To roughly align with the electronic structure, %
the basis functions depend on the molecular structure, hence are denoted as $\{\eta_{\clM,\alpha}(\bfrr)\}_{\alpha=1}^B$.
Expansion coefficients of the orbitals are collected into a matrix $\bfC \in \bbR^{B \times N}$ in the following way:
$\phi_i(\bfrr) = \sum_{\alpha=1}^B \bfC_{\alpha i} \eta_{\clM,\alpha}(\bfrr)$.
DFT typically solves the electronic energy minimization problem w.r.t $\bfC$ by directly solving the optimality equation:
\begin{align}
  \bfH_\clM(\bfC) \, \bfC = \bfS_\clM \, \bfC \, \bfeps,
  \label{eqn:ks-eq-mat}
\end{align}
which is called the Kohn-Sham equation. %
Here, $\bfH_\clM(\bfC)$ is a matrix-valued function with an explicit expression (given an exchange-correlation functional) (\appxref{build-fock}). This matrix is called the Hamiltonian matrix (also noted as the Fock matrix) due to the resemblance of the equation to the Schr\"odinger equation.
The matrix $\bfS_{\clM,\alpha\beta} := \int \eta_{\clM,\alpha}(\bfrr) \eta_{\clM,\beta}(\bfrr) \dd \bfrr$ is the overlap matrix of the basis, which can be computed analytically for common basis choices.
\eqnref{ks-eq-mat} can be seen as a generalized eigenvalue problem defined by the matrices $\bfH_\clM(\bfC)$ and $\bfS_\clM$, where the coefficients of orbitals $\bfC$ in the equation can be understood as eigenvectors, and the diagonal matrix $\bfeps$ comprises eigenvalues which are referred to as orbital energies.

However, the difficulty to solve \eqnref{ks-eq-mat} is that, the matrix that defines the problem and the eigenvector solution are intertwined: the eigenvectors $\bfC$ need to recover the Hamiltonian matrix that defined the eigenvalue problem through the explicit function $\bfH_\clM(\bfC)$ (\figref{sc-loss}, middle). %
Conventional DFT calculation solves it using a fixed-point iteration process called self-consistent field (SCF) iteration.
In each step, orbital coefficients $\bfC^{(k-1)}$ are used to construct the Hamiltonian matrix $\bfH^{(k)} := \bfH_\clM(\bfC^{(k-1)})$, which defines a generalized eigenvalue problem $\bfH^{(k)} \bfC = \bfS_\clM \bfC \bfeps$, whose %
eigenvectors, denoted as $\bfC_\clM(\bfH^{(k)})$, are taken as the updated orbital coefficients $\bfC^{(k)}$ (\figref{sc-loss}, top right).
The converged Hamiltonian $\bfH_\clM^\star$ and its eigenvectors hence solve \eqnref{ks-eq-mat}, which then derive various molecular structures.

Hamiltonian prediction aims to bypass the SCF iteration by directly predicting $\bfH_\clM^\star$ from molecular structure $\clM$ using a machine-learning model $\bfHh_\theta(\clM)$,\footnote{The ``hat'' or ``circumflex'' accent in the notation here is meant to represent ``a neural-network estimator''. %
}
where $\theta$ denotes the model parameters to be learned.
The conventional way to learn such a model is by supervised learning, which requires running DFT on a set of molecular structures $\clD$ to construct a labeled dataset $\overline{\clD}$,
on which the supervised training loss function is applied: %
\begin{align}
  \clL_\data(\theta; \overline{\clD}) := \frac1{|\overline{\clD}|} \sum_{(\clM,\bfH_\clM^\star) \in \overline{\clD}} \lrVert{\bfHh_\theta(\clM) - \bfH_\clM^\star}_\rmF^2,
  \label{eqn:data-loss}
\end{align}
where $|\overline{\clD}|$ denotes the number of samples in the set $\overline{\clD}$. The squared Frobenius norm amounts to the mean squared error (MSE) over the matrix entries. Some works~\citep{unke2021se,yu2023efficient} also include a mean absolute error (MAE) loss for more efficient learning.

\subsection{Self-Consistency Training} \label{sec:mth-sc}

We now describe the proposed self-consistency training for Hamiltonian prediction.
It can be seen as another way to solve the Kohn-Sham equation~\eqref{eqn:ks-eq-mat}, which the prediction $\bfHh_\theta(\clM)$ needs to satisfy.
Recall that the equation is equivalent to the condition that $\bfC := \bfC_\clM(\bfH)$, \ie, the eigenvectors of the generalized eigenvalue problem defined by the Hamiltonian matrix $\bfH$, construct the same Hamiltonian matrix, \ie, $\bfH_\clM(\bfC) = \bfH$ (\figref{sc-loss}, middle).
The self-consistency training loss is hence designed to enforce this condition: the difference between the predicted Hamiltonian $\bfHh_\theta(\clM)$ and the reconstructed Hamiltonian from itself should be minimized, where the reconstruction is done by first solving for the eigenvectors $\bfC_{\clM,\theta} := \bfC_\clM\big( \bfHh_\theta(\clM) \big)$ of the generalized eigenvalue problem defined by $\bfHh_\theta(\clM)$ then constructing the Hamiltonian using $\bfH_\clM(\bfC_{\clM,\theta})$ (\figref{sc-loss}, bottom right).
Explicitly, the self-consistency loss is:
\begin{align}
  &\clL_\SC(\theta; \clD) := \\
  &\frac1{|\clD|} \sum_{\clM \in \clD} \Big\Vert \bfHh_\theta(\clM) - \bfH_\clM\Big(\bfC_\clM \big( \bfHh_\theta(\clM) \big) \Big) \Big\Vert_\rmF^2.
  \label{eqn:sc-loss}
\end{align}
Following the practice of previous work~\citep{unke2021se,yu2023efficient}, we also include its MAE counterpart in place of the squared Frobenius norm into the loss. %
The implementation process is summarized in \algref{sc-loss}.
Note that the loss only requires a set of molecular structures $\clD$ unnecessarily with Hamiltonian labels. It thus enables leveraging numerous molecular structures for learning Hamiltonian prediction, which could substantially enhance generalizability of the prediction model to a wide range of molecules, allowing applicability beyond the limitation of labeled datasets.

We make the following four remarks regarding the understanding of the self-consistency loss.
\itemone~The self-consistency loss is distinct from regularization or self-supervised training, in the sense that the loss by itself can already drive the model towards the exact target, since the loss enforces the equation that determines the target. %
\itemtwo~We emphasize that the loss should \emph{not} be interpreted as updating the prediction $\bfHh_\theta(\clM)$ towards the reconstructed Hamiltonian $\bfH_\clM\big(\bfC_\clM \big( \bfHh_\theta(\clM) \big) \big)$ as a fixed target (which is the case when the \texttt{stop\_grad} operation is applied to the reconstructed Hamiltonian), and the back-propagation (\ie, computation of the gradient of the loss w.r.t $\theta$) through the Hamiltonian reconstruction process is indispensable.
This is because the reconstructed Hamiltonian unnecessarily comes closer to the target solution~\citep{pulay1982improved,cances2000convergence}, %
so taking the reconstructed Hamiltonian as a constant when optimizing $\theta$ may even make the model worse.
Instead, the loss aims to minimize the change from the reconstruction process. To minimize this change, both the predicted matrix and the reconstructed matrix are driven towards the solution.
\itemthr~One may also consider enforcing self-consistency by minimizing the difference in the derived energy after reconstruction, which meets the common stopping criterion in a DFT calculation and could hold more physical relevance. But this would require eigen-solving the reconstructed matrix and evaluating the energy from the eigenvectors, which is as costly as another Hamiltonian reconstruction, making the loss unacceptably costly to optimize.
\itemfor~The self-consistency loss bears some similarity to the SCF loss in DM21~\citep{kirkpatrick2021pushing}. Both connect a DFT solution and an exchange-correlation (XC) functional defining the DFT calculation (part of $\bfH_\clM(\bfC)$ in our formulation). The SCF loss is used to regularize an XC functional model with a label (solution), while we use the self-consistency loss to train a solution-prediction model given a well-established XC functional. It is future work to investigate the utility of the SCF loss for unsupervised Hamiltonian prediction.

\subsection{Implementation Considerations}
\label{sec:mth-detail}

For stable and efficient optimization of the self-consistency loss, we mention a few technical treatments.

\paragraph{Back-Propagation through Eigensolver.}
As mentioned, back-propagation through the reconstruction process $\bfH_\clM\big(\bfC_\clM \big( \bfHh_\theta(\clM) \big) \big)$ is indispensable. This requires differentiation through the eigensolver $\bfC_\clM(\bfH)$.
We leverage the eigensolver implemented in an automatic differentiation package PyTorch~\cite{paszke2019pytorch} which automatically provides the differentiation calculation.
Nevertheless, the calculation often appears numerically unstable~\citep{ionescu2015matrix,wang2019backpropagation}, as it relies on a matrix $\bfG$ (see \appxref{exp-set-sc} for detailed derivation),
\begin{align}
    \bfG_{ij} = 
    \begin{cases}
        1/(\epsilon_i - \epsilon_j), &i \neq j, \\
        0, &i = j,
    \end{cases}
\end{align}%
where $\epsilon_i$ is $i$-th eigenvalue. When there are two close eigenvalues, the values in $\bfG$ can be exceedingly large, %
causing unstable training.
To mitigate this instability, we introduce two treatments.
The first is simply truncating the values in $\bfG$ if they are larger than a chosen threshold.
The second treatment is to skip the model parameter update when the scale of the gradient w.r.t parameters exceeds a certain threshold. Appendix~\ref{appx:exp-set-sc} presents more details.

\begin{algorithm}[t]
\caption{Implementation of self-consistency loss (on one molecular structure)} 
\label{alg:sc-loss}
\begin{algorithmic}[1]
\REQUIRE Molecular structure $\clM = \{\clZ, \clR\}$ comprising types $\clZ := \{Z^{(a)}\}_{a=1}^A$ and coordinates $\clR := \{\bfR^{(a)}\}_{a=1}^A$ of its atoms, pre-computed integral matrices (\eg, overlap matrix $\bfS_{\clM}$), Hamiltonian prediction model $\bfHh_\theta(\cdot)$ to be learned.
\STATE Generate requisite integrals and quadrature grid for constructing Hamiltonian (\appxref{exp-set-sc});
\STATE Predict Hamiltonian $\bfHh_\theta(\clM)$ using the model;
\STATE Solve for the eigenvectors $\bfC_{\clM,\theta}$ of the generalized eigenvalue problem $\bfHh_\theta(\clM) \, \bfC = \bfS_\clM \, \bfC \, \bfeps$.
\STATE Reconstruct Hamiltonian $\bfH_\clM(\bfC_{\clM,\theta})$ following an explicit expression (Appendix~\ref{appx:build-fock});
\STATE Compute the loss $\clL_\SC(\theta; \{\clM\})$ as the addition of the mean squared error (shown in \eqnref{sc-loss}) and mean absolute error between $\bfHh_\theta(\clM)$ and $\bfH_\clM(\bfC_{\clM,\theta})$.
\OUTPUT $\clL_\SC(\theta; \{\clM\})$.
\end{algorithmic}
\end{algorithm}

\paragraph{Efficient Hamiltonian Reconstruction.}
Evaluating the function $\bfH_\clM(\bfC)$ is also a costly procedure, mainly due to two computational components.
The first is the evaluation of basis functions on a quadrature grid for evaluating the exchange-correlation component of the Hamiltonian matrix (\appxref{build-fock}). To accelerate this part, we implemented a GPU-based evaluation process of basis functions on grid points.
The other costly procedure is the evaluation of the Hartree component of the Hamiltonian matrix, which requires $O(N^4)$ cost in its vanilla form. For efficient evaluation of this term, we adopt the density fitting approach (\appxref{build-fock}), a widely used technique in DFT to reduce the complexity to $O(N^3)$ with acceptable loss of accuracy.

\begin{table*}[ht]
    \vspace{-0.1in}
    \caption{Generalization improvement by self-consistency training on unlabeled data in the \emph{data-scarce} scenario (MD17 Hamiltonian).
    Evaluated on the test split of conformations of each molecule.
    }
    \centering
    \small
    \setlength\tabcolsep{4pt}
    \begin{tabular}{lccccccccc}
    \toprule
    Molecule & Setting & $\bfH\,[\mu E_\mathrm{h}]\downarrow$ & $\bfeps\,[\mu E_\mathrm{h}]\downarrow$ & $\bfC\,[\%]\uparrow$ & $\epsilon_{\mathrm{HOMO}}$$\,[\mu E_\mathrm{h}]\downarrow$ & $\epsilon_{\mathrm{LUMO}}$$\,[\mu E_\mathrm{h}]\downarrow$ & $\epsilon_\Delta$$\,[\mu E_\mathrm{h}]\downarrow$ & SCF Accel.$\,[\%]\downarrow$ \\
    \midrule
    \multirow{2}{*}{Ethanol}  & label & 160.36  & 712.54  & 99.44 & 911.64 & 6800.84 & 6643.11 & 68.3 \\
            & label\,+\,\scloss & \textbf{75.65} & \textbf{285.49} & \textbf{99.94} & \textbf{336.97} & \textbf{1203.60} & \textbf{1224.86} & \textbf{61.5} \\
    \midrule
    Malondi- & label  & 101.19 & 456.75 & 99.09 & 471.92& 1093.22  & 1115.94& 69.1 \\
    aldehyde  & label\,+\,\scloss & \textbf{86.60} & \textbf{280.39} & \textbf{99.67} & \textbf{274.45} & \textbf{279.14} & \textbf{324.37} & \textbf{62.1} \\
    \midrule
    \multirow{2}{*}{Uracil} & label & 88.26  & 1079.51 & 95.83 & 1217.17 & 12496.1 & 11850.56 & 65.8 \\
           & label\,+\,\scloss & \textbf{63.82} & \textbf{315.40} & \textbf{99.58} & \textbf{359.98} & \textbf{369.67} & \textbf{388.30} & \textbf{54.5} \\
    \bottomrule
    \end{tabular}
    \label{tab:res-md17}
    \vspace{-0.1in}
\end{table*}

\subsection{Amortization of DFT Calculation} \label{sec:mth-amort}

As mentioned in \secref{mth-sc}, self-consistency training can be applied to unlimited unlabeled molecular structures, hence can substantially improve the generalizability of a Hamiltonian prediction model.
Here, we point out that self-consistency training is also more efficient to improve generalizability than generating additional labels using DFT on those data and then supervising the model.
This is based on the interpretation that self-consistency training is an \emph{amortization} of DFT:
for a given molecular structure $\clM$, DFT requires \emph{multiple} SCF iterations for convergence before it can provide a supervision on $\clM$ (\figref{sc-loss}, top right), while self-consistency training only requires \emph{one} SCF iteration to evaluate the loss and guide the training on $\clM$ (\figref{sc-loss}, bottom right).
This indicates that given the same amount of computational resources measured in the number of SCF iterations, self-consistency training can distribute the resource on more molecular structures, hence providing information on a larger range of the input space.
This is more valuable than Hamiltonian labels on fewer molecular structures for the model to generalize to a broad range of molecular structures.

Self-consistency training can also be viewed as a way to carry out DFT calculation.
Under this view, the amortization effect makes self-consistency training a more efficient method than the conventional DFT to solve a large amount of molecular structures.
Apart from the amortization effect, the efficiency is also benefited from the generalization of a Hamiltonian prediction model to similar molecular structures, on which the model can already provide close results.
The demand to solve a set of molecular structures is not uncommon; \eg, high-throughput drug screening requires investigating a large amount of ligand-receptor compounds using DFT~\citep{jordaan2020virtual}.
Therefore, the applicability scope of Hamiltonian prediction with self-consistency training is enlarged.

\vspace{-1pt}
\section{Experimental Results}
\label{sec:exp}
\vspace{-1pt}

\begin{table*}[htb]
    \vspace{-0.07in}
    \caption{Generalization improvement by self-consistency training on unlabeled data in the \emph{OOD} scenario (QH9).
    The model is trained on the QH9-small training split, and evaluated on the QH9-large test split directly (\texttt{zero-shot}) or after fine-tuned by \texttt{self-con}sistency loss on QH9-large training split (without labels).
    }
    \centering
    \small
    \begin{tabular}{lcccccccc}
    \toprule
    Setting & $\bfH\,[\mu E_\mathrm{h}]\downarrow$ & $\bfeps\,[\mu E_\mathrm{h}]\downarrow$ & $\bfC\,[\%]\uparrow$ & $\eps_{\mathrm{HOMO}}$$\,[\mu E_\mathrm{h}]\downarrow$ & $\eps_{\mathrm{LUMO}}$$\,[\mu E_\mathrm{h}]\downarrow$ & $\eps_\Delta$$\,[\mu E_\mathrm{h}]\downarrow$ & SCF Accel.$\,[\%]\downarrow$ \\
    \midrule
    zero-shot & 69.67 & 403.52 & 95.72 & 778.86 & 12230.49 & 12203.12 & 66.3\\
    \scloss\,(all-param) & 65.74 & 375.31 & \textbf{97.31} & 565.50 & \textbf{1130.55} & \textbf{1316.96} & \textbf{64.5} \\
    \scloss\,(adapter) & \textbf{64.48} & \textbf{268.83} & 97.12 & \textbf{449.80} & 1220.54 & 1394.29 & 65.0 \\
    \bottomrule
    \end{tabular}
    \label{tab:res-qh9}
    \vspace{-0.1in}
\end{table*}

\begin{figure*}[htb]
    \centering
    \includegraphics[width=0.8\linewidth]{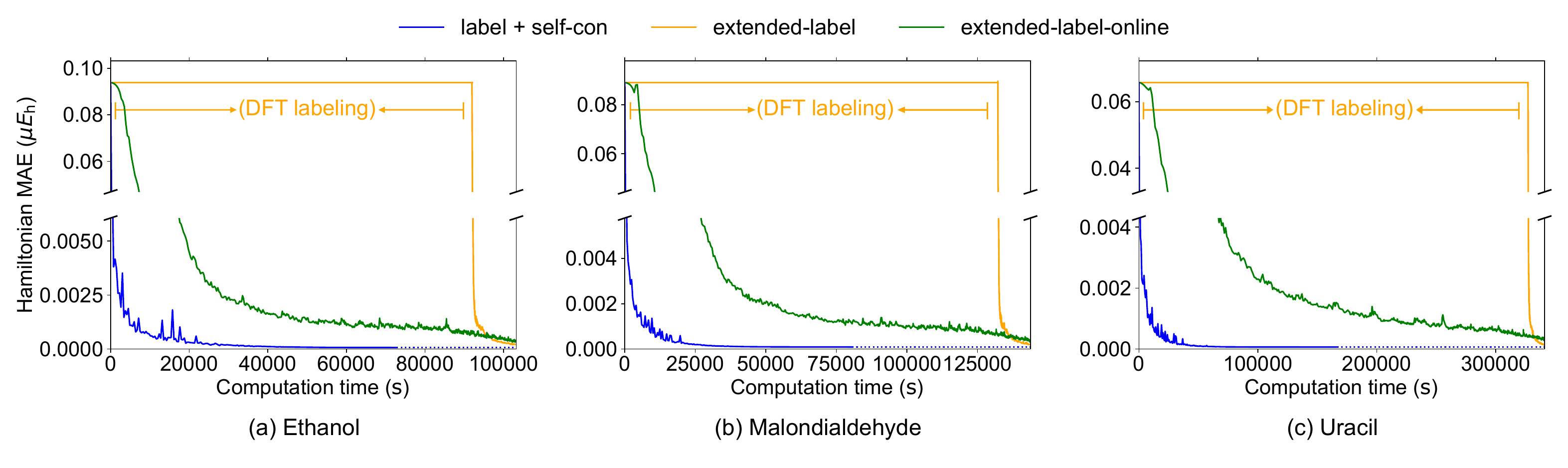}
    \vspace{-0.2in}
    \caption{Efficiency comparison in the \emph{data-scarce} scenario (MD17 Hamiltonian) among \texttt{self-con}sistency training on unlabeled data, supervised training following DFT labeling on unlabeled data (\texttt{extended-label}), and supervised training along with DFT labeling (\texttt{extended-label-online}).
    Dotted horizontal lines extend from the last measured point of the respective curves.
    }
    \label{fig:amor-md17}
    \vspace{-0.1in}
\end{figure*}

\begin{figure*}[htb]
    \centering
    \includegraphics[width=0.6\linewidth]{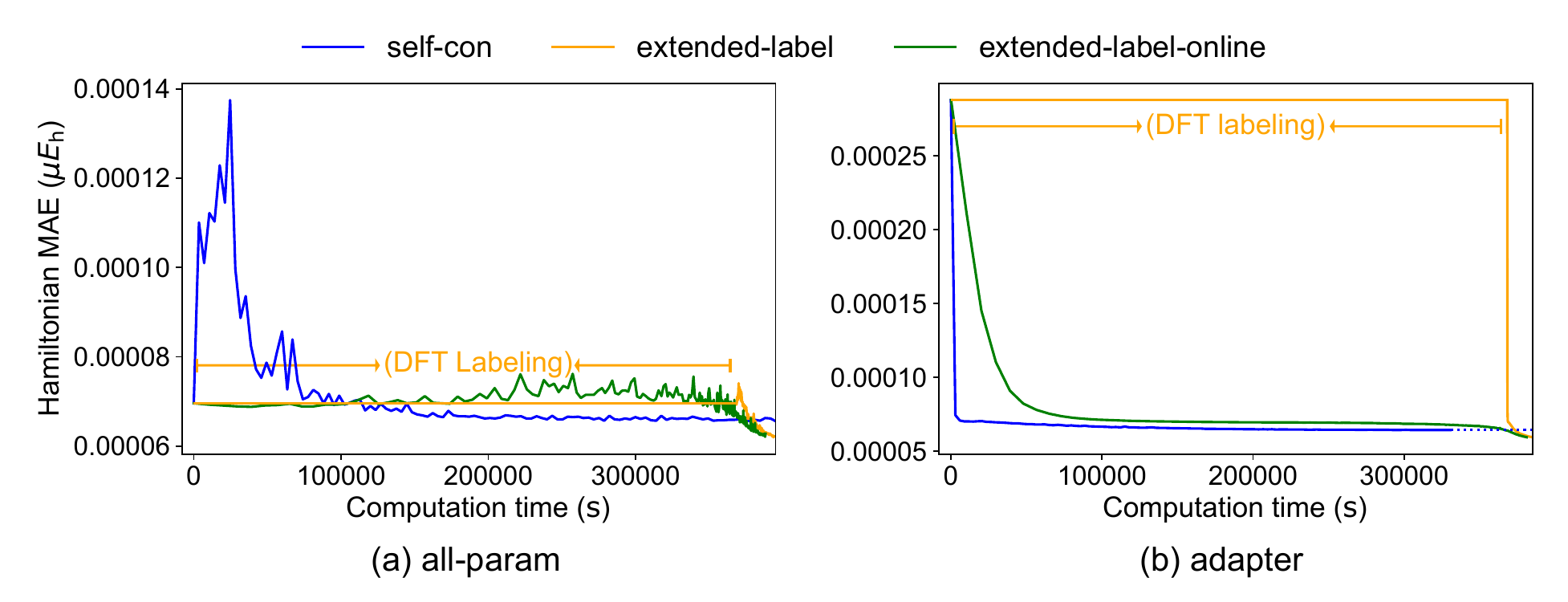}
    \vspace{-0.15in}
    \caption{Efficiency comparison in the \emph{OOD} scenario (QH9) among fine-tuning using \texttt{self-con}sistency training on unlabeled data, supervised training following DFT labeling on unlabeled data (\texttt{extended-label}), and supervised training along with DFT labeling (\texttt{extended-label-online}).
    Dotted horizontal lines extend from the last measured point of the respective curves.
    }
    \label{fig:amor-qh9}
    \vspace{-0.10in}
\end{figure*}

\begin{table*}[htb]
    \caption{Performance comparison between self-consistency training, and supervised training using \emph{full extended labels}, in the \emph{OOD} scenario, corresponding to the ending points of \figref{amor-qh9} (\texttt{extended-label-online} is close to \texttt{extended-label}).
    }
    \centering
    \small
    \setlength\tabcolsep{4pt}
    \begin{tabular}{lccccccccc}
    \toprule
    FT mode & Setting & $\bfH\,[\mu E_\mathrm{h}]\downarrow$ & $\bfeps\,[\mu E_\mathrm{h}]\downarrow$ & $\bfC\,[\%]\uparrow$ & $\eps_{\mathrm{HOMO}}$$\,[\mu E_\mathrm{h}]\downarrow$ & $\eps_{\mathrm{LUMO}}$$\,[\mu E_\mathrm{h}]\downarrow$ & $\eps_\Delta$$\,[\mu E_\mathrm{h}]\downarrow$ & SCF Accel.$\,[\%]\downarrow$ \\
    \midrule
    \multirow{2}{*}{all-param} & extended-label & \textbf{62.13} & \textbf{365.66} & 96.89 & 577.46 & 5962.16 & 6137.66 & 65.0 \\
     & \scloss & 65.74 & 375.31 & \textbf{97.31} & \textbf{565.50} & \textbf{1130.55} & \textbf{1316.96} & \textbf{64.5} \\
     \midrule
    \multirow{2}{*}{adapter} & extended-label & \textbf{59.67} & 330.05 & 96.63 & 541.92 & 6372.12 & 6445.33 & 65.2 \\
    & \scloss & 64.48 & \textbf{268.83} & \textbf{97.12} & \textbf{449.80} & \textbf{1220.54} & \textbf{1394.29} & \textbf{65.0} \\
    \bottomrule
    \end{tabular}
    \label{tab:res-cross}
    \vspace{-0.1in}
\end{table*}

We now empirically validate the benefits of self-consistency training.
We adopt QHNet~\citep{yu2023efficient} as the Hamiltonian prediction model, which is an $\mathrm{SE(3)}$-equivariant graph neural network that balances efficiency and accuracy. Additional results based on alternative architectures (\eg, PhiSNet~\citep{unke2021se}) are provided in Appendix~\ref{appx:exp-alt-archi}, which indicate the same conclusions as presented below.

We employ the following metrics to measure prediction accuracy.
A direct metric is the mean absolute error (MAE) over matrix entries between the predicted and DFT-solved Hamiltonian matrices, as introduced by \citet{schutt2019unifying}.
Directly derived quantities from Hamiltonian, including orbital energies $\bfeps$ and coefficients $\bfC$ solved from the generalized eigenvalue problem, are also used to assess accuracy, measured by MAE for $\bfeps$ and cosine similarity for $\bfC$.
We also report the MAE for three molecular properties %
relevant to molecular research,
including the highest occupied molecular orbital energy $\eps_{\mathrm{HOMO}}$, the lowest unoccupied molecular orbital energy $\eps_{\mathrm{LUMO}}$, and their gap $\eps_\Delta$. %
We also assess the utility for accelerating DFT by the ratio of the number of SCF steps to convergence using the prediction as initialization over the number using the standard initialization, denoted as ``SCF Accel.''
The conventional %
DIIS~\citep{pulay1980convergence} strategy is adopted for running SCF iteration, while we also present ``SCF Accel.'' results using the second-order SCF (SOSCF)~\citep{sun2017general} iteration strategy in Appendix~\ref{appx:exp-res-init}, considering that DIIS may lead to non-monotone iterations thereby diminishes the benefit of a more accurate initialization.

\vspace{-1pt}
\subsection{Self-Consistency Training Improves Generalization}
\label{sec:exp-gen}
\vspace{-1pt}

As discussed in \secref{mth-sc}, self-consistency training can leverage unlabeled data to improve generalizability. We validate this benefit on two challenging generalization scenarios.

\vspace{-3pt}
\paragraph{Data-Scarce Scenario.}
For some scientific tasks with limited labels available, it is difficult for the machine learning model to achieve meaningful performance even for in-distribution (ID) generalization. To demonstrate the advantage of self-consistency training in this scenario, we first conduct generalization experiments over the conformational space. Conformations of ethanol, malondialdehyde and uracil from the MD17 dataset~\citep{chmiela2019sgdml,schutt2019unifying} are considered. The training/validation/test split setting follows \citet{schutt2019unifying}. To simulate a data-scarce setting, for each molecule, only 100 labeled conformations (denoted as $\overline{\clD^{(1)}}$) are provided for supervised training using the supervised loss $\clL_\data(\theta; \overline{\clD^{(1)}})$ (\eqnref{data-loss}).
With the self-consistency loss (\eqnref{sc-loss}), a large amount of additional unlabeled structures in the training set (about 24,900 structures for each molecule; denoted as $\clD^{(2)}$) can be leveraged,
in which case the resulting loss function is: %
{\setlength\abovedisplayskip{4pt}
\setlength\belowdisplayskip{4pt}
\begin{align}\label{eqn:tot-loss}
  \clL_\data(\theta; \overline{\clD^{(1)}}) + \lambda_\SC \, \clL_\SC(\theta; \clD^{(2)}).
\end{align}}%
See more training details in Appendix~\ref{appx:exp-set-train}.

Prediction results on test structures are summarized in \tabref{res-md17}. Compared to the results of supervised (label-based) training, applying self-consistency loss on unlabeled structures leads to a substantial improvement across all evaluation metrics and molecules. Notably, it achieves a significant reduction in the Hamiltonian MAE, with decreases from 14.4\% to 52.8\%. The MAE of $\eps_{\mathrm{HOMO}}$, $\eps_{\mathrm{LUMO}}$ and $\eps_\Delta$ are even reduced by several folds. Applying self-consistency training also substantially improves the acceleration for conventional DFT. In addition, a point-by-point, instance-level comparison in Appendix~\ref{appx:exp-res-accel} shows that self-consistency training leads to faster SCF convergence over various molecular systems consistently, while supervised training does not.
These findings underscore the attractive capability of self-consistency training in breaking the limitation of data scarcity.

\paragraph{Out-of-Distribution (OOD) Scenario.}
\citet{yu2023qh9} introduced the QH9 dataset to benchmark Hamiltonian prediction over the chemical space. Their findings highlight a challenging out-of-distribution (OOD) generalization scenario: models trained on smaller molecules often struggle to generalize to larger molecules, restricting the applicability.
To demonstrate the effect of better generalization using self-consistency training, we construct a similar OOD scenario. We split the molecular structures in QH9 into two subsets: QH9-small comprising molecules with no more than 20 atoms, and QH9-large with larger molecules. The two subsets are then correspondingly divided at random into distinct training/validation and training/validation/test splits (see more dataset details in Appendix~\ref{appx:exp-set-data}).
The model is trained and validated on QH9-small using the supervised loss (\eqnref{data-loss}), and is tested on the QH9-large test split (dubbed \texttt{zero-shot}).
With the self-consistency loss (\eqnref{sc-loss}), the model is allowed to be fine-tuned (without the pretraining supervised loss) on relevant but unlabeled molecular structures, for which we take the QH9-large training split.
We consider two fine-tuning settings: fine-tuning all parameters of the model, dubbed \texttt{self-con (all-param)}, or introducing an adapter module atop the model which is the only optimized component, dubbed \texttt{self-con (adapter)}. We ensure all models are sufficiently trained. Appendix~\ref{appx:exp-set-train} shows more training details.

From the results shown in \tabref{res-qh9}, we observe a significant improvement by fine-tuning using self-consistency on unlabeled QH9-large molecules in both fine-tuning settings. Remarkably, self-consistency reduces the MAE of $\eps_{\mathrm{LUMO}}$ and $\eps_\Delta$ by an order of magnitude.
This result demonstrates that self-consistency training enables the flexibility to adapt a model to an OOD workload without labeled data.

\subsection{Self-Consistency Training is More Efficient than DFT Labeling}
\label{sec:exp-amor}

As discussed in \secref{mth-amort}, self-consistency training can train a model more efficiently than DFT labeling followed by supervised learning, due to its amortization effect of DFT calculation. We demonstrate the empirical efficiency by comparing self-consistency training/fine-tuning in the above two scenarios to the alternative approach of generating labels by running DFT on the additional unlabeled molecular structures %
then applying supervised training using these extended labels (dubbed \texttt{extended-label}).
We also consider a variant that conducts DFT labeling along with model training (dubbed \texttt{extended-label-online}): DFT is only run on unlabeled molecular structures in the current training batch drawn at random, and the generated labels are stored for possible use in future batches. This could be more efficient than \texttt{extended-label}.

The efficiency is monitored by the accuracy-cost curve along training. The accuracy is measured by the validation Hamiltonian MAE.
The cost can be measured by \emph{the number of SCF steps} along self-consistency training or DFT labeling. This matches the analysis in \secref{mth-amort} and is system- and implementation-independent. Results are shown in Figs.~\ref{fig:amor-md17-effscf} and~\ref{fig:amor-qh9-effscf} in Appendix~\ref{appx:exp-res-amor}, which validates the better efficiency in all cases.

For better practical relevance and considering the complication of the interplay between running SCF and model parameter optimization, we present results measured by the cost of real \emph{computation time} here.
All methods are implemented on a workstation equipped with an NVIDIA A100 GPU with 80 GiB memory and a 24-core AMD EPYC CPU.
\paragraph{Data-Scarce Scenario.}
Accuracy-cost curves of the three training strategies are presented in \figref{amor-md17}. We see that self-consistency training converges rapidly, achieving a low prediction error with a cheap cost.
In contrast, the \texttt{extended-label} strategy keeps a plateau at first, representing the process to generate labels using DFT during which the model is not optimized. It is only after the DFT labeling process that the prediction error starts to drop.
The \texttt{extended-label-online} strategy indeed improves upon \texttt{extended-label} by amortizing the labeling cost over the course of training, but it is still not as efficient as self-consistency training, whose amortization capability allows a more frequent model optimization per SCF step.
We note that due to our hardware limitation, DFT labeling and model optimization are performed sequentially. The two processes can be  parallelized which may further improve the efficiency of \texttt{extended-label-online} at the cost of using more machines.
\paragraph{Out-of-Distribution (OOD) Scenario.}
All the three training settings are run for the fine-tuning stage of the model. Curves on QH9-large validation split are presented in \figref{amor-qh9}.
We see again that self-consistency training achieves a high accuracy at a relatively low cost across both fine-tuning settings. In contrast, \texttt{extended-label} and \texttt{extended-label-online} require a higher computational cost to reach a comparable level of accuracy.
These results indicate the better efficiency of the self-consistency training through the amortization effect.

\paragraph{Performance of Final Results.}
At the end of the accuracy-cost curves in \figref{amor-qh9}, computational resource is sufficient to generate full extended labels for supervised training in the OOD scenario, which has the most abundant and direct supervision information, hence serves as an upper bound of Hamiltonian prediction performance.
But as shown in \tabref{res-cross}, this only applies to the Hamiltonian MAE, corresponding to the directly supervised quantity, while self-consistency training still excels at derived physical quantities, especially on $\eps_{\mathrm{HOMO}}$, $\eps_{\mathrm{LUMO}}$ and $\eps_\Delta$, which are directly concerned molecular properties thus more relevant to practical applications.
Appendix~\ref{appx:exp-res-gen} shows a similar observation in the data-scarce scenario.
This indicates that even with unlimited computational resource, self-consistency training could still be preferred than generating labels to better support real applications.

\paragraph{Direct Acceleration over DFT Calculation.}
As mentioned at the end of \secref{mth-amort}, self-consistency training is also a way to directly accelerate DFT calculation on a large amount of molecular structures by leveraging its amortization effect.
To demonstrate the advantage empirically, we compare the computation time of self-consistency training (using \eqnref{tot-loss}) $t_\SC$ and of DFT calculation $t_\textnormal{DFT}$ for solving the unlabeled molecular structures in the data-scarce scenario (see \secref{exp-gen}).
The same stopping criterion is applied to both methods in each case, which is taken as the error of electronic energy (\eqnref{engopt-coeff}) derived from the Hamiltonian following the convention in DFT calculation.
Results are shown in \tabref{time-amor}. We see that self-consistency training indeed requires lower computational cost than DFT to reach the same level of accuracy, demonstrating the practical benefit of amortization.
Appendix~\ref{appx:exp-res-amor} shows more implementation details.

\begin{table}[t]
    \vspace{-0.1in}
    \caption{Efficiency comparison between self-consistency training and \emph{conventional DFT} for solving MD17 molecular structures. Computation times under the same stopping criteria are shown for solving the unlabeled molecular structures in the data-scarce scenario.}
    \centering
    \small
    \begin{tabular}{l@{\hspace{-0pt}}ccc}
    \toprule
     Molecule & criterion $[\mu E_\mathrm{h}]$ & $t_\textnormal{self-con}\,[\mathrm{s}]$ & $t_\textnormal{DFT}\,[\mathrm{s}]$ \\
     \midrule
     Ethanol  & 31.0 & \textbf{4.50}$\bm{\e{4}}$ & 6.40$\e{4}$ \\
     Malondialdehyde  &  88.9 & \textbf{4.81}$\bm{\e{4}}$ & 1.05$\e{5}$ \\
     Uracil  &  177.2 & \textbf{1.23}$\bm{\e{5}}$ & 2.15$\e{5}$ \\
    \bottomrule
    \end{tabular}
    \label{tab:time-amor}
    \vspace{-0.05in}
\end{table}

\begin{table*}[ht]
    \caption{Generalization to \emph{larger-scale} molecules (from MD22) than previously reported in Hamiltonian prediction, with comparison to generalization results of \texttt{e2e} property predictors.
    Models are pretrained on QH9-full with labels and directly evaluated on the molecules (\texttt{zero-shot}, \texttt{e2e}), or, for the Hamiltonian model, after fine-tuned by \texttt{self-con}sistency without labels. %
    Self-consistency training enables meaningful prediction on the larger molecules by bridging generalization gap, and significantly outperforms \texttt{e2e} predictors in molecular properties.
    }
    \centering
    \small
    \setlength\tabcolsep{4pt}
    \begin{tabular}{lccccccccc}
    \toprule
    Molecule & Setting & $\bfH\,[\mu E_\mathrm{h}]\downarrow$ & $\bfeps\,[\mu E_\mathrm{h}]\downarrow$ & $\bfC\,[\%]\uparrow$ & $\eps_{\mathrm{HOMO}}$$\,[\mu E_\mathrm{h}]\downarrow$ & $\eps_{\mathrm{LUMO}}$$\,[\mu E_\mathrm{h}]\downarrow$ & $\eps_\Delta$$\,[\mu E_\mathrm{h}]\downarrow$ & SCF Accel.$\,[\%]\downarrow$ \\
    \midrule
    \multirow{4}{*}{ALA3} & zero-shot & 237.71 & 6.54$\times10^{3}$ & 52.24 & 6.90$\times10^{3}$ & 9.51$\times10^{4}$ & 9.79$\times10^{4}$ & 84.6\\
     & \scloss & \textbf{52.49} & \textbf{1.22$\times10^{3}$} & \textbf{94.46} & \textbf{2.07$\times10^{3}$} & \textbf{3.76$\times10^{3}$} & \textbf{2.69$\times10^{3}$} & \textbf{64.7}\\
     \cmidrule{2-9}
      & e2e (ET) & N/A & N/A& N/A & 1.74$\times10^{5}$ & 7.72$\times10^{3}$ & 2.38$\times10^{5}$ & N/A \\
      & e2e (Equiformer) & N/A & N/A& N/A & 2.38$\times10^{5}$ & 1.16$\times10^{4}$ & 2.27$\times10^{5}$ & N/A \\
     \midrule
    \multirow{4}{*}{DHA} & zero-shot & 397.87 & 1.84$\times10^{4}$ & 20.15 & 1.11$\times10^{4}$ & 1.90$\times10^{5}$ & 1.85$\times10^{5}$ & 170.8\\
     & \scloss & \textbf{56.12} & \textbf{1.81$\times10^{3}$} & \textbf{83.51} & \textbf{1.99$\times10^{3}$} & \textbf{4.01$\times10^{3}$} & \textbf{2.34$\times10^{3}$} & \textbf{67.0} \\
     \cmidrule{2-9}
      & e2e (ET) & N/A & N/A& N/A & 2.92$\times10^{5}$ & 2.58$\times10^{4}$ & 3.39$\times10^{5}$ & N/A \\
      & e2e (Equiformer) & N/A & N/A& N/A & 3.76$\times10^{5}$ & 2.31$\times10^{4}$ & 4.17$\times10^{5}$ & N/A \\
    \bottomrule
    \end{tabular}
    \label{tab:res-md22}
    \vspace{-0.1in}
\end{table*}

\subsection{Self-Consistency Training Extends the Scale of Hamiltonian Prediction}
\label{sec:exp-large}

After verifying the advantages of self-consistency training, we now wield this powerful tool to extend Hamiltonian prediction to molecules larger than previously reported in the field, hence enhancing the relevance to real applications.
\paragraph{Extension to a Larger Scale.}
For molecules larger than those covered by Hamiltonian prediction previously, we consider two molecules in the MD22 dataset~\citep{chmiela2023accurate}: Ac-Ala3-NHMe (ALA3, 42 atoms) and DHA (56 atoms). Both molecules exceed the size of the largest molecule in the QH9 dataset (31 atoms) with a significant gap.
Due to the extensive DFT cost, we generate labels for each molecule only on 500 randomly selected structures from MD22, which are used only for evaluation.
The model is pre-trained on nearly all QH9 molecules (the QH9-full setting in \tabref{data-qh9}), then got \texttt{all-param} fine-tuned using the self-consistency loss (\eqnref{sc-loss}) on the selected structures but without using the labels. %
For ease of training, we employ the MINAO initialization~\citep{sun2018pyscf} as a base Hamiltonian and let the model predict the residual correction.

The results are presented in \tabref{res-md22}.
Compared to the previously best setting \texttt{zero-shot} which can only be directly used right after pre-trained on QH9, fine-tuning with self-consistency substantially improves the performance on the two large molecules, with the MAE of $\eps_{\mathrm{LUMO}}$ and $\eps_\Delta$ reduced by an order, and at least 3x less for other properties.
We note the inadequate performance of \texttt{zero-shot} is not due to insufficient training on QH9, since the validation Hamiltonian MAE of 29.06$\,\mu E_\mathrm{h}$ is sufficiently low. The performance gap is %
due to the substantial scale gap between QH9 and MD22. This gap indicates generalization to larger-scale molecules is highly challenging.
Remarkably, %
self-consistency training breaks the data limitation and achieves a significant performance improvement. %
Moreover, when considering the acceleration benefit for SCF, \texttt{zero-shot} prediction only brings a limited acceleration %
and even results in deceleration (on DHA). %
In contrast, self-consistency training consistently achieves more significant SCF acceleration. %
Appendix~\ref{appx:exp-res-init} presents SCF acceleration results under the SOSCF iteration strategy, where the improvement by self-consistency training is even more significant.
\paragraph{Comparison to End-to-End Property Predictors.}
The conventional paradigm for molecular property prediction is to predict each property with a devoted model in an end-to-end (\texttt{e2e}) manner~\citep{schutt2018schnet,gasteiger2020directional,tholke2021equivariant,liao2022equiformer}, while Hamiltonian prediction offers the advantage to provide all properties that DFT can provide using a single model. Moreover,
self-consistency training differentiates Hamiltonian prediction from other property prediction tasks in that it enables continued improvement of Hamiltonian prediction without labeled data. This continued improvement can even spread to various molecular properties, which cannot be achieved by \texttt{e2e} predictors without additional labeled data.

We showcase this unique merit in this scenario of generalizing to larger-scale molecules, where the continued improvement could bridge the generalization gap. We compare the properties $\eps_{\mathrm{HOMO}}$, $\eps_{\mathrm{LUMO}}$ and $\eps_\Delta$ derived from the predicted Hamiltonian with the direct prediction results by the respective \texttt{e2e} predictors.
Same as the Hamiltonian predictor, the \texttt{e2e} property predictors are trained with labels on nearly all QH9 molecules, but they do not have a self-consistency training strategy hence can only be directly applied to predict the respective properties on the much larger molecules.
We consider two representative implementations of the \texttt{e2e} predictors, using the ET~\cite{tholke2021equivariant} and the Equiformer~\cite{liao2022equiformer} architectures. %

Results are shown in \tabref{res-md22}.
They indicate that the \texttt{e2e} predictors significantly suffer from the generalization gap when comparing the evaluated error to the validation error (\tabref{appx-res-md22}) which are around 1$\e{3}\,\mu E_\mathrm{h}$.
The Hamiltonian predictor can also predict such properties as derived from the predicted Hamiltonian. Even applied right after pre-training (\texttt{zero-shot}), the Hamiltonian predictor can already provide better results than \texttt{e2e} predictors on $\eps_{\mathrm{HOMO}}$ and $\eps_\Delta$.
Using the unique lever of self-consistency training, the Hamiltonian predictor provides much more accurate results on all three properties, with one to two orders less MAE than \texttt{e2e} predictors.
These results indicate that self-consistency training offers a promising avenue towards improving OOD prediction for molecular properties in a label-free manner.

\vspace{-.2cm}
\section{Conclusion} \label{sec:concl}

We have presented self-consistency training for Hamiltonian prediction, a novel training method that does not require labeled data. This is a unique advantage of Hamiltonian prediction. As the self-consistency loss is designed to enforce %
the basic equation of DFT, it provides complete and exact information of the prediction target. Self-consistency training opens the access to gain supervision from vastly available unlabeled data, which substantially solves the data-scarce problem and allows generalization to challenging domains. We have also pointed out and empirically verified that self-consistency training is more efficient than running DFT to generate data for supervised learning, benefited from its amortization effect. Using self-consistency training, we have pushed Hamiltonian prediction to solve molecules larger than ever reported.

More broadly, since Hamiltonian matrix can derive rich molecular properties (\eg, energy, HOMO-LUMO gap), the self-consistency training can also improve the prediction of these properties without labeled data, and can even supervise end-to-end prediction models. Since labeled data in the science domain in general is much less than in conventional AI domains, and generalization is more challenging due to the flexibility in the input, this way to leverage fundamental physical laws to train the model would be especially helpful.

\section*{Acknowledgements}
We thank Lin Huang, Han Yang, and Yue Wang for insightful discussions on the idea and techniques; Erpai Luo for discussions on model design and help with dataset preparation; Jan Hermann, Michael Gastegger and Sebastian Ehlert for suggestions on evaluation and writing; Tao Qin, Jia Zhang and Huanhuan Xia for constructive feedback. Additionally, we acknowledge Lin Huang, Han Yang, and Jia Zhang for providing their CuDA implementation of Hamiltonian construction. We also thank anonymous reviewers and area chair for their feedback.
He Zhang and Nanning Zheng were supported by the National Natural Science Foundation of China under Grant 62088102.

\section*{Impact Statement}

This paper presents work whose goal is to advance the field of 
Machine Learning. There are many potential societal consequences 
of our work, none which we feel must be specifically highlighted here.

\bibliography{main}
\bibliographystyle{icml2024}

\newpage
\appendix
\onecolumn
\numberwithin{equation}{section}
\numberwithin{figure}{section}
\numberwithin{table}{section}

\newcommand{\bfLt}{\tilde{\bfL}}
\newcommand{\bfWt}{\tilde{\bfW}}
\newcommand{\bfGammab}{\bar{\bfGamma}}
\newcommand{\flatten}{\mathrm{vec}}

\section{Brief Introduction to Density Functional Theory} \label{appx:dft}

\subsection{Background of Electronic Structure Methods}

All properties of a molecule is determined by the result of interaction among the electrons and nuclei in the molecule.
As nuclei are much heavier than electrons, %
they are typically treated as classical particles, while the electrons are governed by the Schr\"odinger equation. %
Therefore, the state of the $A$ nuclei is specified by the molecular structure $\clM := \{\clR, \clZ\}$, where $\clZ := \{Z^{(a)}\}_{a=1}^A$ and $\clR := \{\bfR^{(a)}\}_{a=1}^A$ specifies the atomic numbers (species) and coordinates of the nuclei, while the state of the $N$ electrons is specified by the wavefunction $\psi(\bfrr^{(1)}, \cdots, \bfrr^{(N)})$. The squared modulus of the wavefunction $\lrvert{\psi(\bfrr^{(1)}, \cdots, \bfrr^{(N)})}^2$ represents the joint distribution of the $N$ electrons. Since electrons %
are indistinguishable, the density function $\lrvert{\psi(\bfrr^{(1)}, \cdots, \bfrr^{(N)})}^2$ is permutation symmetric, hence the wavefunction $\psi(\bfrr^{(1)}, \cdots, \bfrr^{(N)})$ is permutation symmetric or antisymmetric, \ie, it keeps or changes sign (\ie, phase change of 0 or $\pi$) when the coordinates of two particles are exchanged. For electrons, their statistical behavior indicates that their wavefunction is antisymmetric (electrons are fermions).

Commonly, only the stationary states of electrons in a given molecular structure $\clM$ are concerned, since the evolution of electrons is much faster than the motion of nuclei so their state instantly becomes stationary for any given molecular structure. The stationary states are determined by the stationary Schr\"odinger equation:
$\clHh_\clM \psi = E \psi$, \ie, they are eigenstates of the Hamiltonian operator $\clHh_\clM$.
The Hamiltonian operator $\clHh_\clM := \Th + \Vh_\ee + \Vh_{\ext,\clM}$ is composed of the kinetic energy operator $\Th \psi := -\frac{1}{2} \sum_{i=1}^N \nabla_{\bfrr^{(i)}}^2 \psi$ (atomic units are used throughout), the internal potential energy operator among electrons $\Vh_\ee \psi := \sum_{1 \le i < j \le N} \frac{1}{\lrVert*{\bfrr^{(i)} - \bfrr^{(j)}}} \psi$, and the external potential energy operator $\Vh_{\ext,\clM} \psi := \sum_{i=1}^N V_{\ext,\clM}(\bfrr^{(i)}) \psi$ where $V_{\ext,\clM}(\bfrr) := -\sum_{a=1}^A \frac{Z^{(a)}}{\lrVert{\bfrr - \bfR^{(a)}}}$ is the external potential generated by the nuclei.
Note that the latter two operators are multiplicative, \ie, their action on a wavefunction is the multiplication with the corresponding potential function.
The Hamiltonian operator is Hermitian, hence its eigenvalues are always real.
Moreover, since this Hamiltonian operator is also real (in physics term, it is time-reversible) (particularly, it does not involve magnetic fields or spin-orbital coupling), every eigenstate of it has a real-valued eigenfunction. Hence from now on, it suffices to only consider real-valued wavefunctions.

Solving an eigenvalue problem is challenging especially when $N$ is large. On the other hand, most of the concerned properties of a molecule only involve the ground state, \ie, the eigenstate with the lowest eigenvalue (energy). Hence an alternative form to solve the electronic ground state can be composed as an optimization problem, known as the variational formulation:
\begin{align}
  E^\star_\clM = \min_{\psi: \, \text{antisym}, \braket{\psi}{\psi} = 1} \obraket*{\psi}{\clHh_\clM}{\psi},
  \label{eqn:engopt-wavefn}
\end{align}
where $\braket{\psi}{\phi}$ denotes the integral of $\psi^* \phi$ w.r.t all their arguments, and $\obraket*{\psi}{\clHh_\clM}{\phi} := \braket*{\psi}{\clHh_\clM \phi}$ (which is also $\braket*{\clHh_\clM \psi}{\phi}$ since $\clHh_\clM$ is Hermitian). Various ways to parameterize the wavefunction $\psi$ and estimate and optimize the energy are proposed. Regardless, this formulation optimizes a function on $\bbR^{3N}$, whose complexity may increase exponentially w.r.t system size $N$, limiting the scale of practically applicable systems.

Before going on, we note that although wavefunctions are in general complex-valued, it is sufficient to only consider real-valued wavefunctions for solving static (\ie, no time evolution) electronic state without magnetic fields and ignoring spin-orbital coupling,
since the Hamiltonian operator $\clHh_\clM$ in such cases are not only Hermitian but also time-reversible (meaning that $\clHh_\clM$ is ``real'', $\clHh_\clM^* = \clHh_\clM$), each of whose eigenstate has a real-valued eigenfunction.
For this reason, we only consider real-valued functions (including orbitals and basis functions), and do not distinguish matrix transpose and Hermitian conjugate.

\subsection{Basic Idea of DFT}

Density functional theory (DFT) is motivated to address the exponentially complex optimization space. It aims to optimize the (one-electron reduced) density $\rho(\bfrr)$, a function on a fixed-dimensional space $\bbR^3$. It is a reduced quantity to describe the electronic state. The electron density corresponding to the electronic state specified by wavefunction $\psi$ is the marginal distribution (up to a factor of the number of electrons) of the joint distribution:
\begin{align}
  \rho_{[\psi]}(\bfrr) := N \int \lrvert*{\psi(\bfrr, \bfrr^{(2)}, \cdots, \bfrr^{(N)})}^2 \dd \bfrr^{(2)} \cdots \ud \bfrr^{(N)},
  \label{eqn:den-def}
\end{align}
which is independent of the variable for which the marginalization is conducted due to the indistinguishability.
It is how one straightforwardly perceives electron density, which is a valid concept also under the classical view.

Now the question is, whether optimizing the density is sufficient to determine the electronic ground state, considering that the density is only a reduced quantity. This is first answered affirmatively in the seminal work by \citet{hohenberg1964inhomogeneous}, but it would be more explicit to deduce the answer following Levy's constrained search formulation~\citep{levy1979universal} of \eqnref{engopt-wavefn}:
\begin{align}
  & E^\star_\clM = \min_{\psi: \, \text{antisym}, \braket{\psi}{\psi} = N} \obraket*{\psi}{\clHh_\clM}{\psi} ={} \min_{\rho: \ge 0, \braket{\bbone}{\rho} = N} \lrparen*[\Big]{ \min_{\psi: \, \text{antisym}, \rho_{[\psi]} = \rho} \obraket*{\psi}{\clHh_\clM}{\psi} },
  \label{eqn:engopt-twolevel}
\end{align}
where $\bbone$ denotes the constant 1-valued function.
Note that when viewing $\min_{\psi: \, \text{antisym}, \rho_{[\psi]} = \rho} \obraket*{\psi}{\clHh_\clM}{\psi}$ as a functional of density $\rho$, the optimization problem in \eqnref{engopt-twolevel} indicates that the ground-state energy and density can indeed be solved by optimizing the density.
Among the three components of $\clHh_\clM$, the $\Vh_{\ext,\clM}$ term already makes a density functional, since $\obraket*{\psi}{\Vh_{\ext,\clM}}{\psi}
= \sum_{i=1}^N \int V_{\ext,\clM}(\bfrr^{(i)}) \lrvert{\psi(\bfrr^{(1)}, \cdots, \bfrr^{(N)})}^2 \dd \bfrr^{(1)} \cdots \ud \bfrr^{(N)}
= \frac1N \sum_{i=1}^N \int V_{\ext,\clM}(\bfrr^{(i)}) \rho_{[\psi]}(\bfrr^{(i)}) \dd \bfrr^{(i)}
= \braket*{V_{\ext,\clM}}{\rho_{[\psi]}}$ is independent of $\psi$ once $\rho_{[\psi]}$ is fixed. So \eqnref{engopt-twolevel} can be formulated as: 
\begin{align}
  E^\star_\clM = \min_{\rho: \ge 0, \braket{\bbone}{\rho} = N} \underbrace{\lrparen*[\Big]{ \min_{\psi: \, \text{antisym}, \rho_{[\psi]} = \rho} \! \obraket*{\psi}{\Th + \Vh_\ee}{\psi} }}_{=: F[\rho]} + \braket*{V_{\ext,\clM}}{\rho},
  \label{eqn:univ-fnal-def}
\end{align}
where $F[\rho]$ is called the universal functional comprising the kinetic and internal potential energy minimally attainable for the given density $\rho$. Its name follows the fact that it does not depend on molecular structure $\clM$ and applies to any system.

As the universal functional is still quite implicit to carry out practical calculation, approximations are considered to cover the major part of the kinetic energy and of the internal potential energy.
For the latter, the classical internal potential energy can be used, which ignores electron correlation and adopts an explicit expression in terms of $\rho(\bfrr)$:
\begin{align}
  E_\tnH[\rho] := \frac{1}{2} \int \frac{\rho(\bfrr) \rho(\bfrr')}{\lrVert{\bfrr - \bfrr'}} \dd \bfrr \ud \bfrr'.
  \label{eqn:hart-def}
\end{align}
It is also called the Hartree energy, hence the notation.
For the kinetic part, the kinetic energy density functional (KEDF) is introduced following a similar formulation as the definition of the universal functional:
\begin{align}
  T_\tnS[\rho] := \min_{\psi: \, \text{antisym}, \rho_{[\psi]} = \rho} \obraket*{\psi}{\Th}{\psi}.
  \label{eqn:ts-def}
\end{align}
The rest part of the kinetic and internal potential energy is called the exchange-correlation (XC) energy:
\begin{align}
  E_\XC[\rho] := F[\rho] - T_\tnS[\rho] - E_\tnH[\rho],
\end{align}
and the variational problem to solve the electronic ground state of molecule in structure $\clM$ becomes:
\begin{align}
  E^\star_\clM = \! \min_{\rho: \ge 0, \braket{\bbone}{\rho} = N} \! T_\tnS[\rho] + E_\tnH[\rho] + E_\XC[\rho] + \braket*{V_{\ext,\clM}}{\rho}.
  \label{eqn:engopt-den}
\end{align}
Although the exact expression of $E_\XC$ in terms of $\rho$ is still unknown, it makes only a minor part of the total electronic energy and is more flexible to approximate. Over the past decades, researchers have developed many successful approximations~\citep{becke1993density,stephens1994ab,perdew1996generalized,perdew2008restoring}. Deep learning has also been leveraged for developing an approximation~\citep{dick2020machine,chen2021deepks,kirkpatrick2021pushing}.
As for the KEDF, there are methods that also directly approximate the density functional~\citep{slater1951simplification,wang1999orbital,witt2018orbital}, which are now called orbital-free density functional theory. Nevertheless, approximating KEDF is harder and requires higher accuracy, since it accounts for a major part of energy. It is also an active research direction to %
leverage machine learning models to approximate the functional more accurately~\citep{snyder2012finding,imoto2021order,remme2023kineticnet,del2023variational,zhang2024overcoming}.

\subsection{Kohn-Sham DFT}

Considering the difficulty of directly approximating the KEDF, \citet{kohn1965self} exploited properties of the KEDF and developed a method that evaluates the kinetic energy directly.
Note that in the optimization of the definition of KEDF in \eqnref{ts-def}, there is no interaction among electrons ($\Th$ operates one-body-wise). It is known that the ground-state wavefunction solution of non-interacting systems is in the form of a determinant (at least in absense of degeneracy~\citep[Thm.~4.6]{lieb1983density}), which, instead of a general function on $\bbR^{3N}$, is composed of $N$ functions $\Phi := \{\phi_i(\bfrr)\}_{i=1}^N$ on $\bbR^3$ called orbitals:
\begin{align}
  \psi_{[\Phi]}(\bfrr^{(1)}, \cdots, \bfrr^{(N)}) := \frac{1}{\sqrt{N!}} \det[\phi_i(\bfrr^{(j)})]_{ij}.
  \label{eqn:slater-det}
\end{align}
The optimization problem in the definition of KEDF in \eqnref{ts-def} can be equivalently formulated as:\footnote{
  Assume the queried density $\rho$ comes from the set of densities of the ground state of all non-interacting systems. Although this set still has $\rho$ that violates the equivalence to \eqnref{ts-def}~\citep[Thm.~4.8]{lieb1983density}, the determinantal definition \eqnref{ts-orb} still recovers the ground-state energy if optimized on this set~\citep[Thm.~4.9]{lieb1983density}.
}
\begin{align}
  T_\tnS[\rho] ={} \min_{\{\phi_i\}_{i=1}^N: \rho_{[\psi_{[\Phi]}]} = \rho} \obraket*{\psi_{[\Phi]}}{\Th}{\psi_{[\Phi]}}
  = \min_{\subalign{\{\phi_i\}_{i=1}^N: & \text{orthonormal}, \\ & \rho_{[\psi_{[\Phi]}]} = \rho}} \sum_{i=1}^N \obraket*{\phi_i}{\Th}{\phi_i},
  \label{eqn:ts-orb}
\end{align}
where the second expression to optimize orthonormal orbitals, $\braket*{\phi_i}{\phi_j} = \delta_{ij}$, is valid since any set of functions can be orthonormalized by \eg, the Gram-Schmidt process without changing the corresponding density and kinetic energy.
This is desired to simplify calculation, for which the kinetic energy calculation is simplified in \eqnref{ts-orb}, and the density (\eqnref{den-def}) substituted by \eqnref{slater-det} can also be simplified as:
\begin{align}
  \rho_{[\psi_{[\Phi]}]}(\bfrr) = \sum_{i=1}^N \lrvert{\phi_i(\bfrr)}^2.
  \label{eqn:den-orb-orthon}
\end{align}

Using this simplified formulation \eqnref{ts-orb}, the original optimization problem \eqnref{engopt-den} for solving the electronic structure given molecular structure $\clM$ becomes: 
\begin{align}
  E^\star_\clM ={} & \!\! \min_{\rho: \ge 0, \braket{\bbone}{\rho} = N}
  \bigg\{ \lrparen*[\bigg]{ \min_{\subalign{\{\phi_i\}_{i=1}^N: & \text{orthonormal}, \\ & \rho_{[\psi_{[\Phi]}]} = \rho}} \sum_{i=1}^N \obraket*{\phi_i}{\Th}{\phi_i} } + E_\tnH[\rho] + E_\XC[\rho] + \braket*{V_{\ext,\clM}}{\rho} \bigg\} \\
  ={} & \!\!\! \min_{\subalign{\{\phi_i\}_{i=1}^N: \\ \text{orthonormal}}} \! \bigg\{ E_\clM[\Phi] := \sum_{i=1}^N \obraket*{\phi_i}{\Th}{\phi_i} + E_\tnH[\rho_{[\psi_{[\Phi]}]}] + E_\XC[\rho_{[\psi_{[\Phi]}]}] + \braket*{V_{\ext,\clM}}{\rho_{[\psi_{[\Phi]}]}} \bigg\}.
  \label{eqn:engopt-orb}
\end{align}
In this way, the query for directly evaluating $T_\tnS[\rho]$ is avoided by an exact estimation using the orbitals. This formulation optimizes $N$ functions on $\bbR^3$ instead of one function on $\bbR^3$, hence the complexity is increased by at least an order of $N$.
This formulation is called the Kohn-Sham DFT, and has become the default DFT formulation due to its success to solve molecular system problems computationally~\citep{seminario1996recent,jain2013materials}.

To solve \eqnref{engopt-orb}, standard DFT solves the equation of optimality, which is derived by taking the variation of $E_\clM[\Phi]$ w.r.t each orbital under the orthonormality constraint.
The variation of $E_\clM[\Phi]$ is: 
\begin{align}
  \fracdelta{E_\clM[\Phi]}{\phi_i}(\bfrr) ={} & \fracdelta{\sum_{j=1}^N (-\frac12) \obraket*{\phi_j}{\nabla^2}{\phi_j}}{\phi_i}(\bfrr) +{} \int \fracdelta{\big( E_\tnH[\rho] + E_\XC[\rho] + \braket*{V_{\ext,\clM}}{\rho} \big)}{\rho(\bfrr')} \Big\vert_{\rho = \rho_{[\psi_{[\Phi]}]}} \!\!\! \fracdelta{\rho_{[\psi_{[\Phi]}]}(\bfrr')}{\phi_i(\bfrr)} \dd \bfrr' \\
  ={} & 2 \Th \phi_i(\bfrr) + 2 \Bigg(\! \underbrace{ \int \frac{\rho(\bfrr')}{\lrVert{\bfrr' - \bfrr}} \dd \bfrr' }_{=: V_{\tnH[\rho]}(\bfrr)}
  + \underbrace{ \fracdelta{E_\XC[\rho]}{\rho}(\bfrr) }_{=: V_{\XC[\rho]}(\bfrr)} \!\Bigg) \!\Bigg\vert_{\rho = \rho_{[\psi_{[\Phi]}]}} \!\!\! \phi_i(\bfrr){}+ 2 V_{\ext,\clM}(\bfrr) \phi_i(\bfrr).
  \label{eqn:eng-orb-variation}
\end{align}
By introducing the (one-electron effective) Hamiltonian operator, or more commonly called the Fock operator in DFT,
\begin{align}
  \Hh_{\clM,[\rho]} := \Th + \Vh_{\tnH[\rho]} + \Vh_{\XC[\rho]} + \Vh_{\ext,\clM},
  \label{eqn:fock-opr}
\end{align}
where the latter three operators act on a function by multiplying the function with the respective potential energy function,
the variation can be written as:
\begin{align}
  \fracdelta{E_\clM[\Phi]}{\phi_i}(\bfrr) = 2 \Hh_{\clM,[\rho_{[\psi_{[\Phi]}]}]} \phi_i.
  \label{eqn:delta-E-orb}
\end{align}
For the orthonormality constraint, first consider the normalization constraint and introduce Lagrange multipliers $\{\veps_i\}_{i=1}^N$ for them. The corresponding variation is:
\begin{align}
  \fracdelta{\sum_{j=1}^N \veps_j (\braket*{\phi_j}{\phi_j} - 1)}{\phi_i}(\bfrr)
  = 2 \veps_i \phi_i(\bfrr),
\end{align}
which leads to the optimality equation:
\begin{align}
  \Hh_{\clM,[\rho_{[\psi_{[\Phi]}]}]} \phi_i(\bfrr) = \veps_i \phi_i(\bfrr), \quad \forall i = 1, \cdots, N.
  \label{eqn:ks-eq}
\end{align}
This is known as the Kohn-Sham equation (in function form).
From this equation, the optimal solution of orbitals are eigenstates of the operator $\Hh_{\clM,[\rho_{[\psi_{[\Phi]}]}]}$, which can be verified to be Hermitian. Hence, in the general case where there is no degeneracy, different orbitals in the solution are naturally orthogonal, so there is no need to further enforce this constraint explicitly.

\subsection{Practical Calculation under a Basis}

Vectorizing a function as the expansion coefficient vector on a basis function set is an effective and controllable way to represent a function numerically.
For molecules, as the electrons distribute around atoms in the molecule, commonly adopted basis functions are atom-centered functions. To allow analytical calculation of integrals, the functions typically take a Gaussian form for the radial variable (\ie, the distance from the center nucleus of this basis function) multiplied with a spherical harmonic function for the angular variables (or equivalently a monomial of the three coordinates)~\cite{ditchfield1971self,hellweg2015development,dunning1989gaussian,jensen2001polarization}.
Different chemical elements usually have different sets of basis functions. To expand the orbitals in a molecule, the basis set is the union of basis functions centered at each of the atoms in the molecule. We collectively label them with one index $\alpha$, and denote them as $\{\eta_{\clM,\alpha}(\bfrr)\}_{\alpha=1}^B$. The number of basis functions $B$ for a molecular system typically increases linearly with the number of electrons $N$ in the system.

The orbitals can then be represented as expansion coefficients $\bfC$:
\begin{align}
  \phi_i(\bfrr) = \sum_{\alpha=1}^B \bfC_{\alpha i} \, \eta_{\clM,\alpha}(\bfrr).
  \label{eqn:orb-expd}
\end{align}
Next we show the derivation for the optimality equation for $\bfC$.
Given that orthonormality constraint is satisfied, the density corresponding to the orbital state specified by $\bfC$ is:
\begin{align}
  \rho_{\clM,\bfC}(\bfrr) &= \sum_{\alpha,\beta} \sum_{i=1}^N \bfC_{\alpha i} \, \bfC_{\beta i} \, \eta_{\clM,\alpha}(\bfrr) \, \eta_{\clM,\beta}(\bfrr) = \sum_{\alpha,\beta} (\bfC \bfC\trs)_{\alpha\beta} \, \eta_{\clM,\alpha}(\bfrr) \, \eta_{\clM,\beta}(\bfrr).
  \label{eqn:den-orbcoeff}
\end{align}
The Kohn-Sham equation presented in \eqnref{ks-eq} is turned into
$\sum_\alpha \bfC_{\alpha i} \Hh_{\clM,[\rho_{\clM,\bfC}]} \eta_{\clM,\alpha}(\bfrr) = \sum_\alpha \veps_i \bfC_{\alpha i} \eta_{\clM,\alpha}(\bfrr)$.
Integrating both sides with basis function $\eta_{\clM,\beta}(\bfrr)$ gives:
\begin{align}
  \bfH_\clM(\bfC) \, \bfC = \bfS \, \bfC \, \bfveps,
  \label{eqn:roothaan-eq}
\end{align}
where:
\begin{align}
  \big(\bfH_\clM(\bfC)\big)_{\alpha\beta} := \obraket*{\eta_{\clM,\alpha}}{\Hh_{\clM,[\rho_{\clM,\bfC}]}}{\eta_{\clM,\beta}},
  \label{eqn:ham-mat=ham-opr}
\end{align}
is the Hamiltonian matrix ($\Hh_{\clM,[\rho_{\clM,\bfC}]}$ defined in \eqnref{fock-opr}),
\begin{align}
  (\bfS_\clM)_{\alpha\beta} := \braket*{\eta_{\clM,\alpha}}{\eta_{\clM,\beta}},
\end{align}
is the overlap matrix of the atomic basis, and
\begin{align}
  \bfveps := \Diag(\veps_1, \cdots, \veps_N),
\end{align}
is a diagonal matrix comprising the eigenvalues.
This is the matrix form of the Kohn-Sham equation \eqnref{ks-eq}, as presented in \eqnref{ks-eq-mat} in the main paper.

To solve \eqnref{roothaan-eq}, conventional DFT calculation uses a fixed-point iteration process known as the self-consistent field (SCF) iteration.
At each iteration step $k$, the last orbital solution $\bfC^{(k-1)}$ is used to construct the Hamiltonian matrix $\bfH^{(k)} := \bfH_\clM(\bfC^{(k-1)})$, and the updated orbital solution $\bfC^{(k)}$ for this step is derived by solving $\bfH^{(k)} \bfC = \bfS \bfC \bfveps$.
There are variants that accelerate the iteration, \eg, the direct inversion in the iterative subspace (DIIS) method~\citep{pulay1982improved,kudin2002blackbox}, which constructs $\bfH^{(k)}$ not only using $\bfH_\clM(\bfC^{(k-1)})$ but also using Hamiltonian matrices from previous steps.

In contrast, our self-consistency approach (\eqnref{sc-loss}) solves \eqnref{roothaan-eq} directly, by minimizing the violation of the equality in terms of the Hamiltonian matrix, $\lrVert{\bfHh_\theta(\clM) - \bfH_\clM\Big(\bfC_\clM\big( \bfHh_\theta(\clM) \big) \Big)}^2_\tnF$, where $\bfC_\clM\big( \bfHh_\theta(\clM) \big)$ denotes the orbital coefficients solved from $\bfHh_\theta(\clM) \bfC = \bfS_\clM \bfC \bfveps$.

\subsection{Details to Construct the Hamiltonian Matrix}
\label{appx:build-fock}

The definition of the Hamiltonian matrix is given by \eqnref{ham-mat=ham-opr} as the product of the Hamiltonian operator on basis functions.
The operator is in turn defined by \eqnref{fock-opr} and \eqnref{eng-orb-variation}, following which the Hamiltonian matrix can be computed from the equations below:
\begin{align}
  \bfH_\clM(\bfC) = \bfT_\clM + \bfV_{\tnH,\clM}(\bfC) + \bfV_{\XC,\clM}(\bfC) + \bfV_{\ext,\clM},
  \label{eqn:hamiltonian}
\end{align}
where:
\begin{align}
  & (\bfT_\clM)_{\alpha\beta} := \obraket*{\eta_{\clM,\alpha}}{\Th}{\eta_{\clM,\beta}}
  = -\frac12 \! \int \! \eta_{\clM,\alpha}(\bfrr) \nabla^2 \eta_{\clM,\beta}(\bfrr) \dd \bfrr, \\
  & (\bfV_{\tnH,\clM}(\bfC))_{\alpha\beta} \!:=\! \obraket*{\eta_{\clM,\alpha}}{V_{\tnH[\rho_{\clM,\bfC}]}}{\eta_{\clM,\beta}}
  \stackrel{\text{\eqnsref{eng-orb-variation,den-orbcoeff}}}{=}\!
  \sum_{\gamma\delta} (\bfDt_\clM)_{\alpha\beta,\gamma\delta} (\bfC \bfC\trs)_{\gamma\delta}, \label{eqn:VH-mat} \\
  & \hspace{1.2cm} \text{where~~} (\bfDt_\clM)_{\alpha\beta,\gamma\delta} \!:=\! \iint \frac{\eta_{\clM,\alpha}(\bfrr) \eta_{\clM,\beta}(\bfrr) \eta_{\clM,\gamma}(\bfrr') \eta_{\clM,\delta}(\bfrr')}{\lrVert{\bfrr - \bfrr'}} \dd \bfrr' \ud \bfrr, \label{eqn:cint-4c2e} \\
  & (\bfV_{\XC,\clM}(\bfC))_{\alpha\beta} \!:=\! \obraket*{\eta_{\clM,\alpha}}{V_{\XC[\rho_{\clM,\bfC}]}}{\eta_{\clM,\beta}} \!=\! \int \! V_{\XC[\rho_{\clM,\bfC}]}\!(\bfrr) \eta_{\clM,\alpha}(\bfrr) \eta_{\clM,\beta}(\bfrr) \dd \bfrr, \label{eqn:VXC-mat} \\
  & (\bfV_{\ext,\clM})_{\alpha\beta} := \obraket*{\eta_{\clM,\alpha}}{V_{\ext,\clM}}{\eta_{\clM,\beta}} = -\sum_{a=1}^A Z^{(a)} \! \int \frac{\eta_{\clM,\alpha}(\bfrr) \eta_{\clM,\beta}(\bfrr)}{\lrVert{\bfrr - \bfR^{(a)}}} \dd \bfrr. \label{eqn:Vext-mat}
\end{align}
Under the mentioned type of basis functions $\{\eta_{\clM,\alpha}(\bfrr)\}_{\alpha=1}^B$, integrals $\bfS_\clM$, $\bfT_\clM$, $\bfDt_\clM$, and $\bfV_{\ext,\clM}$ can be evaluated analytically.
To evaluate $\bfV_{\XC,\clM}(\bfC)$, the integral can be evaluated on a quadrature grid, for which common XC functional approximations provide a way to evaluate $V_{\XC[\rho_{\clM,\bfC}]}$ on each grid point.

\paragraph{Density Fitting}
Note that directly calculating $\bfV_{\tnH,\clM}(\bfC)$ following \eqnref{VH-mat} requires $O(B^4) = O(N^4)$ complexity, which soon dominates the cost and restricts the applicability to large systems. There is a widely adopted approach in DFT to reduce the complexity for this term, called \emph{density fitting}~\citep{dunlap2000robust}.
It is motivated by noting $V_{\tnH[\rho_{\clM,\bfC}]}(\bfrr)$ as defined in \eqnref{eng-orb-variation} involves an integral with density function $\rho_{\clM,\bfC}(\bfrr)$, which, by \eqnref{den-orbcoeff}, involves a double summation that incurs $O(N^2)$ cost.
\eqnref{den-orbcoeff} can be seen as expanding the density function onto the paired basis set $\{\eta_{\clM,\alpha}(\bfrr) \eta_{\clM,\beta}(\bfrr)\}_{\alpha,\beta = 1, \cdots, B}$ of size $B^2$. It is hence possible to reduce the complexity by projecting the density function onto an \emph{auxiliary basis set} $\{\omega_{\clM,\mu}(\bfrr)\}_{\mu=1}^M$ of size $M = O(N)$.
The projected density can be represented by the corresponding coefficients $\bfpp$ in the way that:
\begin{align}
  \rho_{\clM,\bfpp}(\bfrr) := \sum_{\mu=1}^M \bfpp_\mu \omega_{\clM,\mu}(\bfrr).
  \label{eqn:den-aux}
\end{align}
The projection is done by finding the coefficients $\bfpp$ that minimizes the difference from $\rho_{\clM,\bfpp}(\bfrr)$ to $\rho_{\clM,\bfC}(\bfrr)$. Note that the purpose of density fitting here is to reduce cost complexity for calculating $\bfV_{\tnH,\clM}$, the operator matrix corresponding to the Hartree energy defined in \eqnref{hart-def}. Therefore, the difference is preferred to be measured in Hartree energy:
\begin{align}
  E_\tnH[\rho_{\clM,\bfpp} - \rho_{\clM,\bfC}]
  ={} & \iint \frac{\big( \rho_{\clM,\bfpp}(\bfrr) - \rho_{\clM,\bfC}(\bfrr) \big) \big( \rho_{\clM,\bfpp}(\bfrr') - \rho_{\clM,\bfC}(\bfrr') \big)}{\lrVert*{\bfrr - \bfrr'}} \dd \bfrr \ud \bfrr' \\
  ={} & \bfpp\trs \bfWt_\clM \, \bfpp - 2 \bfpp\trs \bfLt_\clM \, \flatten(\bfC \bfC\trs) + \flatten(\bfC \bfC\trs)\trs \bfDt_\clM \, \flatten(\bfC \bfC\trs),
\end{align}
where $\flatten(\bfC \bfC\trs) \in \bbR^{B^2}$ denotes the vector of the flattened density matrix $\bfC \bfC\trs \in \bbR^{B \times B}$,
and the pre-computed constant integral matrices are defined by:
$(\bfWt_\clM)_{\mu\nu} := \iint \frac{\omega_{\clM,\mu}(\bfrr) \omega_{\clM,\nu}(\bfrr')}{\lrVert{\bfrr - \bfrr'}} \dd \bfrr \ud \bfrr'$,
$(\bfLt_\clM)_{\mu,\alpha\beta} := \iint \frac{\omega_{\clM,\mu}(\bfrr) \eta_{\clM,\alpha}(\bfrr') \eta_{\clM,\beta}(\bfrr')}{\lrVert{\bfrr - \bfrr'}} \dd \bfrr \ud \bfrr'$, and
$(\bfDt_\clM)_{\alpha\beta,\gamma\delta} := \iint \frac{\eta_{\clM,\alpha}(\bfrr) \eta_{\clM,\beta}(\bfrr) \eta_{\clM,\gamma}(\bfrr') \eta_{\clM,\delta}(\bfrr')}{\lrVert{\bfrr - \bfrr'}} \dd \bfrr \ud \bfrr'$,
which can be computed analytically using common basis sets.
As a quadratic form, the solution is:
\begin{align}
  \bfpp_\clM(\bfC) := \bfWt_\clM^{-1} \bfLt_\clM \, \flatten(\bfC \bfC\trs).
  \label{eqn:jfit-res}
\end{align}
Note that since the auxiliary basis is usually not complete to expand the paired basis, the projected density $\rho_{\clM,\bfpp_\clM(\bfC)}$ is an approximation to the original density $\rho_{\clM,\bfC}$.

Using density fitting, the Hartree operator matrix $\bfV_{\tnH,\clM}(\bfC)$ defined by \eqnref{VH-mat} can be approximately estimated by substituting the projected density $\rho_{\clM,\bfpp_\clM(\bfC)}(\bfrr)$, given by \eqnref{den-aux} and \eqnref{jfit-res}, into the Hartree potential $V_{\tnH[\rho_{\clM,\bfpp_\clM(\bfC)}]}$ defined by \eqnref{eng-orb-variation}:
\begin{align}
  (\bfV_{\tnH,\clM}(\bfC))_{\alpha\beta} \approx
  \big( \bfpp_\clM(\bfC)\trs \bfLt_\clM \big)_{\alpha\beta}.
  \label{eqn:VH-mat-jfit}
\end{align}
Since the calculation of $\bfpp_\clM(\bfC)$ following \eqnref{jfit-res} has complexity $O(M^3) + O(M B^2) + O(N B^2) = O(N^3)$, and the complexity of \eqnref{VH-mat-jfit} itself has complexity $O(M B^2) = O(N^3)$, the overall complexity for estimating $\bfV_{\tnH,\clM}(\bfC)$ using density fitting is $O(N^3)$, which reduces the original quartic $O(N^4)$ complexity.

\paragraph{Alternative Derivation}
We would like to mention an alternative derivation of the Hamiltonian matrix as an amendment.
This derivation is to first parameterize the optimization problem using a function basis then deriving the optimality condition in matrix form.
Noting that under a basis set $\{\eta_{\clM,\alpha}(\bfrr)\}_{\alpha=1}^B$, the orbital functions can be parameterized using the orbital coefficient matrix $\bfC$ as $\Phi_{\clM,\bfC}$ in the form of \eqnref{orb-expd}, the corresponding optimization problem \eqnref{engopt-orb} can be converted into a usual optimization problem on vectors/matrix (instead of on functions): $E^\star_\clM =$
\begin{align}
  \min_{\substack{\bfC \in \bbR^{B \times N}: \\ \bfC\trs \bfS_\clM \bfC = \bfI}} \! \Bigg\{ \!
  E_\clM(\bfC) := E_\clM[\Phi_{\clM,\bfC}] = \! \Bigg(
  \begin{aligned}    
    & \; \flatten(\bfT_\clM)\trs \flatten(\bfGamma(\bfC))
    + \frac12 \flatten(\bfGamma(\bfC))\trs \bfDt_\clM \flatten(\bfGamma(\bfC)) \\    
    & \!\!+\! E_\XC\big[ \sum_{\alpha,\beta} \bfGamma(\bfC)_{\alpha\beta} \eta_{\clM,\alpha} \eta_{\clM,\beta} \big]
    + \flatten(\bfV_{\ext,\clM})\!\trs  \!\flatten(\bfGamma(\bfC))
  \end{aligned}
  \! \Bigg) \!
  =: E_\clM\big( \bfGamma(\bfC) \big)
  \! \Bigg\},
  \label{eqn:engopt-coeff}
\end{align}
where the constraint comes from the orthonormality of orbitals $\delta_{ij} = \braket*{\phi_{\clM,\bfC,i}}{\phi_{\clM,\bfC,j}} = \sum_{\alpha,\beta} \bfC_{\alpha i} \bfC_{\beta j} \braket*{\eta_{\clM,\alpha}}{\eta_{\clM,\beta}} = \sum_{\alpha,\beta} \bfC_{\alpha i} \bfC_{\beta j} (\bfS_\clM)_{\alpha\beta} = (\bfC\trs \bfS_\clM \bfC)_{ij}$,
and the density matrix is defined by $\bfGamma(\bfC) := \bfC \bfC\trs$.
The expression for the XC energy part comes from the density function expression under a basis, \ie, \eqnref{den-orbcoeff}.
Noting that the energy expression depends on $\bfC$ only through the density matrix $\bfGamma(\bfC)$, we finally denote the optimization objective as $E_\clM\big( \bfGamma(\bfC) \big)$.
Introducing Lagrange multipliers grouped into a symmetric matrix $\bfeps$ (since the constraint is symmetric) for the constraint and taking the gradient w.r.t $\bfC$, we have the optimality condition:
$\nabla_\bfC E_\clM\big( \bfGamma(\bfC) \big) = \nabla_\bfC \tr\big( \bfeps\trs (\bfC\trs \bfS_\clM \bfC - \bfI) \big)$. Using the chain rule and that all matrices except $\bfC$ are symmetric, we have:
\begin{align}
  \nabla_\bfC E_\clM\big(\bfGamma(\bfC)\big) = 2 \nabla_\bfGamma E_\clM\big(\bfGamma(\bfC)\big) \, \bfC,
  \label{eqn:grad-coeff-E}
\end{align}
and that $\nabla_\bfC \tr\big( \bfeps\trs (\bfC\trs \bfS_\clM \bfC - \bfI) \big) = 2 \bfS_\clM \bfC \bfeps$. The optimality equation then becomes:
\begin{align}
  \nabla_\bfGamma E_\clM\big(\bfGamma(\bfC)\big) \, \bfC &= \bfS_\clM \bfC \bfeps.
  \label{eqn:roothaan-eq-grad}
\end{align}
When optimality is achieved, $\bfC$ is the eigenvectors of the Hermitian (symmetric) matrix $\nabla_\bfGamma E_\clM\big(\bfGamma(\bfC)\big)$, so in the common situation that there is no degenerated state, the eigenvectors are already orthogonal, \ie, the non-diagonal part of the constraint $\bfC\trs \bfS_\clM \bfC = \bfI$ is satisfied. Therefore, the multipliers only need to handle the normalization constraints hence only the diagonal part of $\bfeps$ is effective. This reduces $\bfeps$ in \eqnref{roothaan-eq-grad} to a diagonal matrix.
In this way, \eqnref{roothaan-eq-grad} becomes identical to \eqnref{roothaan-eq}, which indicates:
\begin{align}
  \bfH_\clM(\bfC) = \nabla_\bfGamma E_\clM\big(\bfGamma(\bfC)\big).
  \label{eqn:H-mat=grad-denmat-E}
\end{align}

The equivalence to the first definition in \eqnref{ham-mat=ham-opr} together with \eqnref{fock-opr} and \eqnref{eng-orb-variation} can be seen from the relation between variation and gradient: for a general functional $F[\cdot]$ and a general parameterized function $f_\theta(x)$, the relation is:
\begin{align}
  \fracpartial{F[f_\theta]}{\theta} = \int \fracdelta{F[f_\theta]}{f}(x) \fracpartial{f_\theta}{\theta}(x) \dd x.
  \label{eqn:diff=variation}
\end{align}
Using this equation and noting that $(\nabla_\bfC E)_{\alpha i}$ means $\fracpartial{E}{\bfC_{\alpha i}}$ and $E_\clM(\bfC) := E_\clM[\Phi_{\clM,\bfC}]$ from \eqnref{engopt-coeff}, we have:
\begin{align}
  \big(\nabla_\bfC E_\clM(\bfC)\big)_{\alpha i}
  &= \int \sum_{j=1}^N \fracdelta{E_\clM[\Phi_{\clM,\bfC}]}{\phi_{\clM,\bfC,j}}(\bfrr) \big( \nabla_\bfC \phi_{\clM,\bfC,j}(\bfrr) \big)_{\alpha i} \dd \bfrr
  = \int \fracdelta{E_\clM[\Phi_{\clM,\bfC}]}{\phi_{\clM,\bfC,i}}(\bfrr) \big( \nabla_\bfC \phi_{\clM,\bfC,i}(\bfrr) \big)_{\alpha i} \dd \bfrr \\
  &\stackrel{\text{\eqnsref{delta-E-orb,orb-expd}}}{=} 2 \int \Hh_{\clM,[\rho_{\clM,\bfC}]} \phi_{\clM,\bfC,i}(\bfrr) \; \eta_{\clM,\alpha}(\bfrr) \dd \bfrr
  = 2 \int \sum_\beta \bfC_{\beta i} \Hh_{\clM,[\rho_{\clM,\bfC}]} \eta_{\clM,\beta}(\bfrr) \; \eta_{\clM,\alpha}(\bfrr) \dd \bfrr \\
  &= 2 \sum\nolimits_\beta \bfC_{\beta i} \obraket*{\eta_{\clM,\alpha}}{\Hh_{\clM,[\rho_{\clM,\bfC}]}}{\eta_{\clM,\beta}}
  \stackrel{\text{\eqnsref{ham-mat=ham-opr}}}{=} 2 \big( \bfH_\clM(\bfC) \, \bfC \big)_{\alpha i},
\end{align}
which means $\nabla_\bfC E_\clM(\bfC) = 2 \bfH_\clM(\bfC) \, \bfC$.
On the other hand, noting $E_\clM(\bfC) = E_\clM\big(\bfGamma(\bfC)\big)$ from \eqnref{engopt-coeff} and noting \eqnref{grad-coeff-E}, we also have $\nabla_\bfC E_\clM(\bfC) = 2 \nabla_\bfGamma E_\clM\big(\bfGamma(\bfC)\big) \, \bfC$.
This also gives $\bfH_\clM(\bfC) = \nabla_\bfGamma E_\clM\big(\bfGamma(\bfC)\big)$, \ie, \eqnref{H-mat=grad-denmat-E}.
From \eqnref{H-mat=grad-denmat-E}, the detailed construction of the Hamiltonian matrix \eqnref{hamiltonian} to~(\ref{eqn:Vext-mat}) can be recovered using the detailed expressions in \eqnref{engopt-coeff}.

\section{Additional Technical Details}
\label{appx:addi-tech}

In this section, we present additional details regarding the implementation of the Hamiltonian prediction model and the self-consistency loss.

\subsection{Model Implementation Details}
\label{appx:exp-set-model}

\paragraph{QHNet.} We build our model upon the official QHNet codebase\footnote{\href{https://github.com/divelab/AIRS/tree/main/OpenDFT/QHNet}{https://github.com/divelab/AIRS/tree/main/OpenDFT/QHNet}, the code is available under the terms of the GPL-3.0 license}, which is an $\mathrm{SE(3)}$-equiavariant graph neural network for Hamiltonian prediction~\cite{yu2023efficient}. With careful architecture design, QHNet achieves a good balance between inference efficiency and accuracy. Its architecture is composed of four key modules: node-wise interaction, diagonal pair, non-diagonal pair and expansion. Given the atom types $Z$ and positions $\clR$ as inputs, the QHNet model employs five layers of node-wise interaction to extract $\mathrm{SE(3)}$-equivariant atomic features. Subsequently, the features of diagonal/non-diagonal atom pairs are fed into diagonal/non-diagonal pair modules respectively to build pairwise representations $\bfff_{aa}$ (diagonal) and $\bfff_{ab}$ (non-diagonal), where $a$ and $b$ denote the atom index. The expansion module then transforms these pairwise representations into blocks of the Hamiltonian matrix. Further information can be found in the original paper~\cite{yu2023efficient}. For our experimental studies, all models are configured with the default parameters specified for QHNet. The neural network codebase is developed using PyTorch\cite{paszke2019pytorch} and PyTorch-Geometric~\cite{fey2019fast}.

\begin{figure}
    \centering
    \includegraphics[width=0.8\textwidth]{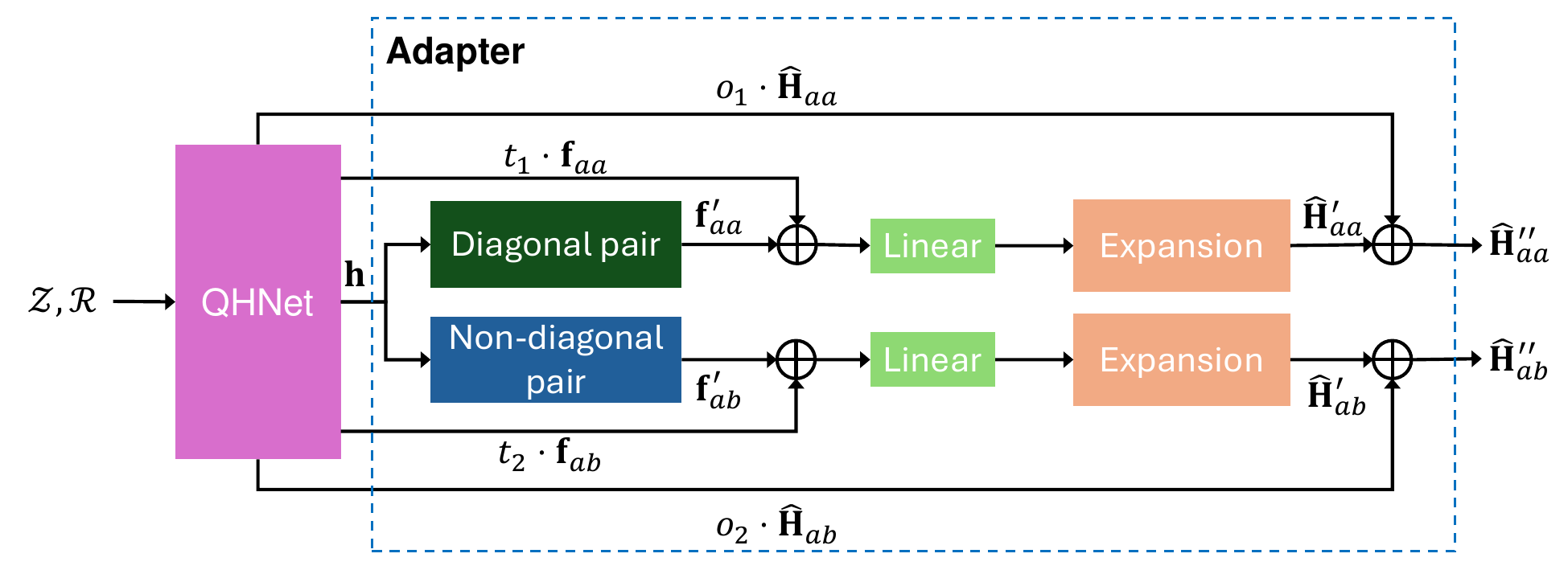}
    \caption{
    The whole architecture of the adapter module.  Given the atom types $\clZ$ and positions $\clR$ as inputs, the pretrained QHNet model is used to produce atomic representations $\bfhh$, pairwise representations $\bfff$ and the initial Hamiltonian prediction $\bfHh$. Subsequently, the adapter module is utilized to produce refinement Hamiltonian $\bfHh^{'}$ based on $\bfhh$ and $\bfff$. Finally, the refinement Hamiltonian is combined with the initial Hamiltonian prediction as the final output $\bfHh^{''}$. $t_1$, $t_2$, $o_1$ and $o_2$ denote learnable combination coefficients. $a$ and $b$ denote the indexes of atoms.}
    \label{fig:adapter}
\end{figure}

\paragraph{Adapter Module. }As outlined in \secref{exp-gen}, to facilitate the generalization of the Hamiltonian model in the OOD scenario, we apply self-consistency loss for fine-tuning the QHNet model with two fine-tuning approaches: \texttt{all-param} and \texttt{adapter}. 
Specifically, we construct the \texttt{adapter} using three modules: diagonal pair module, non-diagonal pair module and expansion module. and then insert it atop the original QHNet model. A schematic illustration of the adapter module is provided in~\figref{adapter}. Given the input molecule, the pretrained QHNet model is initially used to produce the initial Hamiltonian matrix blocks $\bfHh$\footnote{The model-predicted Hamiltonian should be formally denoted as $\bfHh_\theta(\clM)$, and we omit $\theta$ and $\clM$ for brevity}, along with the final atomic representations $\bfhh$ and the final pairwise representations $\bfff$. Subsequently, the atomic representations are fed into corresponding diagonal or non-diagonal pair modules respectively to build pairwise representations $\bfff^{'}$. Afterward, the pairwise representations of the QHNet model and the adapter module are combined (\eg, $\bfff_{aa}^{'} + t_1 \cdot \bfff_{aa}$, with $t_1$ as a learnable combination coefficient).
The combined pairwise representations are first fed into a linear layer and then employed by the expansion module to produce the refinement Hamiltonian ($\bfHh^{'}$).
Finally, we take the combination of the initial Hamiltonian and the refinement Hamiltonian (\eg, $\bfHh_{aa}^{'} + o_1 \cdot \bfHh_{aa}$, with $o_1$ as a learnable combination coefficient) as the final output $\bfHh^{''}$.
It is important to note that the combination of pairwise representations and Hamiltonian blocks, whether diagonal or non-diagonal, is conducted independently, and the combination coefficients are distinct for each pair (\ie, $t_1 \neq t_2$ and $o_1 \neq o_2$).

\subsection{Self-Consistency Loss} \label{appx:exp-set-sc}

\paragraph{Back-Propagation through Eigensolver. }As described in \secref{mth-detail}, the evaluation of self-consistency loss $\clL_\SC$ (\eqnref{sc-loss}) requires the eigenvectors $\bfC_{\clM, \theta}$ of the generalized eigenvalue problem (Line 3 in~\algref{sc-loss}), necessitating the back-propagation through an eigensolver. Thus we need to compute the gradient of the loss function $\clL_\SC$ w.r.t the matrix $\bfHh_\theta(\clM)$. In our practical implementation, we solve the generalized eigenvalue problem for each molecule with two steps: \itemone Solve the eigenvalue problem for matrix $\bfS_\clM$, $\mathbf{\Lambda}, \bfU= \texttt{EigSol}(\bfS_\clM)$ and then define $\bfA=\bfU \mathbf{\Lambda}^{-1/2}$. This leads to the transformation of the Hamiltonian matrix to $\tilde{\bfH}_\theta=\bfA^{\trs}\bfHh_\theta(\clM)\bfA$; \itemtwo Solve the eigenvalue problem for the transformed Hamiltonian matrix $\tilde{\bfH}_\theta$, $\bfeps, \tilde{\bfC}=\texttt{EigSol}(\tilde{\bfH}_\theta)$,  from which the eigenvectors of the original problem are recovered as $\bfC_{\clM, \theta}=\bfA \tilde{\bfC}$. Following these steps, the self-consistency loss $\clL_\SC (\bfH_\clM(\bfC_{\clM, \theta}))$ is calculated using the eigenvectors $\bfC_{\clM, \theta}$ that have been derived.

The partial derivatives of self-consistency loss $\clL_\SC$ w.r.t the transformed matrix $\tilde{\bfH}_\theta$ can be expressed as:
$\nabla_{\tilde{\bfH}_\theta}\clL_\SC=\tilde{\bfC} \big( \bfG \circ (\tilde{\bfC}^{\trs} \nabla_{\tilde{\bfC}}\clL_\SC) \big) \tilde{\bfC}^{\trs}$, where
\begin{align}
    \bfG_{ij} = 
    \begin{cases}
        1/(\epsilon_i - \epsilon_j), &i \neq j, \\
        0, &i = j,
    \end{cases}
\end{align}
with $\epsilon_i$ representing the $i$-th eigenvalues.
The gradient $\nabla_{\tilde{\bfC}}\clL_\SC$ is calculated using the chain rule $\nabla_{\tilde{\bfC}}\clL_\SC = \nabla_{\tilde{\bfC}}\bfC \, \nabla_{\bfC_{\clM, \theta}}\clL_\SC=\mathrm{vec}^{-1} \big((\bfI_N \otimes \bfA^{\trs}) \, \mathrm{vec}(\nabla_{\bfC_{\clM, \theta}}\clL_\SC)\big)$,
where $\mathrm{vec}$ and $\mathrm{vec}^{-1}$ denote the vectorization operator and its inverse  operator, $\otimes$ denotes the Kronecker product operator, and $\bfI_N$ denotes the $N$-dimensional identity matrix.
Then we can derive the partial derivative w.r.t the original Hamiltonian matrix $\bfHh_\theta(\clM)$ as: 
$\nabla_{\bfHh_\theta(\clM)}\clL_\SC = \nabla_{\tilde{\bfH}_\theta}\bfHh_\theta(\clM) \, \nabla_{\tilde{\bfH}_\theta}\clL_\SC=\mathrm{vec}^{-1}\big(\bfA \otimes \bfA \, \mathrm{vec}(\nabla_{\tilde{\bfH}_\theta}\clL_\SC)\big)$. Consequently, the partial derivatives w.r.t $\bfHh_\theta(\clM)$ rely on the matrix $\bfG$, which can lead to a large gradient when two eigenvalues are close.

To mitigate this instability and promote stable training, we introduce two treatments. The first is to limit the magnitude of gradient by applying truncation on matrix $\bfG$ in the \texttt{backward} function of PyTorch:
\begin{align}
  \tilde{\bfG}_{ij} = 
  \begin{cases}
    T\cdot \mathrm{sgn}(\epsilon_i - \epsilon_j), & \text{if } 1/\vert\epsilon_i - \epsilon_j\vert > T, \\
    \bfG_{ij}, & \text{if } 1/\vert\epsilon_i - \epsilon_j\vert \leq T,
  \end{cases}
\end{align}where $T$ is a threshold determined by taking the 60-th percentile of absolute values of all $\bfG$ entries, and $\mathrm{sgn}(\cdot)$ denotes the sign function. The technique is chosen for its simplicity and effectiveness, and there exist other methods for addressing this issue~\cite{song2021approximate,wang2021robust}.
The second treatment is to skip model parameter update when the scale of the gradient w.r.t parameters exceeds a certain threshold $g_s$, which is determined through cross-validation.

\begin{table}[ht]
    \centering
    \caption{Comparison of training time per iteration for the QHNet model with and without the incorporation of the self-consistency loss. The batch size is maintained at 5 across all configurations, and the average training time is calculated over 50 iterations. Unit: $\mathrm{s}$.}
    \begin{tabular}{lccc}
    \toprule
    Model & Ethanol & Malondialdehyde & Uracil \\
    \midrule
    QHNet \emph{without} \scloss & 0.283 & 0.289 & 0.355 \\
    QHNet \emph{with} \scloss & 0.375 & 0.401 & 0.613\\
    \bottomrule
    \end{tabular}
    \label{tab:time-comp}
    \vspace{-0.2in}
\end{table}

\paragraph{Efficient Hamiltonian Reconstruction} To reconstruct the Hamiltonian $\bfH_\clM(\bfC_{\clM,\theta})$, we first generate requisite integrals and quadrature grid using PySCF 
and then compute the Hamiltonian using PyTorch according to standard SCF procedure. Yet,
this involves two costly steps. \itemone 
Evaluating atomic basis functions on generated quadrature grid points is computationally expensive. 
To accelerate this computation, we re-implement the evaluation of basis functions on GPU. Moreover, the grid level determines the number of grid points and in turn influences the construction accuracy of the exchange-correlation potential $\bfV_{\XC,\clM}(\bfC)$. Empirically, we find that a grid level of 2 strikes an optimal balance between construction accuracy and computational efficiency. \itemtwo The computation of the Hartree component entails a $O(N^4)$ complexity (Line~6 in~\algref{sc-loss}, $N$ is the number of electrons). As the molecular size increases, this computation becomes highly costly. To enable an efficient evaluation of the Hartree matrix, we apply the density fitting technique widely used in DFT programs to reduce the computational complexity from $O(N^4)$ to $O(N^3)$.
Leveraging the two techniques leads to a significant acceleration for the Hamiltonian reconstruction, enabling faster self-consistency training. Moreover, integral matrices that are solely dependent on the molecular conformation (\ie, $\bfT_\clM$, $\bfV_{\mathrm{ext}, \clM}$) are pre-computed and stored in the database. These pre-computed matrices are then loaded as needed during the training process.

\paragraph{Computational Complexity. }As the self-consistency loss is constructed following the standard SCF procedure, it possesses the same computational complexity as one SCF iteration under the Kohn-Sham DFT formulation. After the application of the density fitting technique, the computational complexity becomes $O(N^3)$ ($N$ denotes the number of electrons in a molecule). Note that self-consistency training only brings extra computational cost during training, while keeps the same cost for Hamiltonian prediction. The empirical time cost of the Hamiltonian prediction model, both with and without the incorporation of self-consistency loss, are detailed in \tabref{time-comp}.

\section{Experimental Study Settings}
\label{appx:exp-set}

In this section, we provide further data preparation and training details for the empirical study presented in \secref{exp}.  
\subsection{Dataset Preparation}
\label{appx:exp-set-data}

To demonstrate the benefits of self-consistent training, we first conduct experiments on two generalization scenarios, corresponding to two molecular datasets, MD17 and QH9, respectively. Afterward, we evaluate the applicability of the Hamiltonian prediction model on large-scale molecules, for which we adopt the MD22 dataset.

\begin{table}[ht]
    \caption{Statistics of the MD17 dataset~\cite{schutt2019unifying}.}
    \centering
    \begin{tabular}{lccccc}
    \toprule
     Molecule & Train (labeled) & Train (unlabeled) & Validation & Test & Molecular size \\
    \midrule
     Ethanol  &  100 & 24,900 & 500 & 4,500 & 9 \\
     Malondialdehyde  &  100 & 24,900 & 500 & 1,478 & 19 \\
     Uracil  &  100 & 24,900 & 500 & 4,500 & 26 \\
    \bottomrule
    \end{tabular}
    \label{tab:data-md17}
\end{table}

\paragraph{MD17. }To evaluate the benefit of self-consistency training in improving generalization for the data-scarce scenario, we adopt the MD17 dataset~\cite{schutt2019unifying}, and focus on three conformational spaces of ethanol ($\mathrm{C_2 H_5 OH}$), malondialdehyde ($\mathrm{CH_2 (CHO)_2}$) and uracil ($\mathrm{C_4 H_4 N_2 O_2}$). The Hamiltonian matrices in this dataset are calculated with the PBE~\citep{perdew1996generalized} exchange-correlation functional and the Def2SVP Gaussian-type orbital (GTO) basis set.
We follow the split setting used by \citet{schutt2019unifying} to divide the structures of each molecule into training/validation/test sets. Moreover, we randomly select 100 labels from the training set for supervised training, while the remaining training structures are utilized as unlabeled data for self-consistency training. The detailed statistics of three conformational spaces are summarized in \tabref{data-md17}.

\begin{table}[ht]
    \caption{Statistics of the QH9 dataset~\cite{yu2023qh9}.}
    \centering
    \begin{tabular}{lccc}
    \toprule
     Data setting & Training & Validation & Test \\
    \midrule
     QH9-small  &  94,001 & 10,000 &  N/A \\
     QH9-large  &  18,000 & 2,000 & 6,830 \\
     QH9-full  &  124,289 &  6,542 & N/A \\
    \bottomrule
    \end{tabular}
    \label{tab:data-qh9}
\end{table}

\paragraph{QH9. }To evaluate the benefit of self-consistency training in improving generaliation for the out-of-distribution (OOD) scenario, we adopt the QH9 dataset\footnote{The dataset is licensed under a Creative Commons Attribution-NonCommercial-ShareAlike 4.0 International License.}~\cite{yu2023qh9}. This dataset is proposed to benchmark Hamiltonian prediction methods in chemical space, consisting of two subsets: QH-stable and QH-dynamic. Here we adopt the QH-stable subset (dubbed as QH9 hereafter), which consists of 130,831 stable small organic molecules with no more than 9 heavy atoms, as well as their corresponding Hamiltonian matrices. The Hamiltonian matrices are calculated with the B3LYP~\cite{becke1992density} exchange-correlation functional and the Def2SVP GTO basis set. To simulate an OOD benchmark, we divide the QH9 dataset into two subsets by molecular size (QH9-small and QH9-large) and further partition them into training/validation/test sets.
Additionally, we establish a separate split setting for the generalization study on large-scale molecules(referred to as QH9-full). Comprehensive statistics related to these division settings are detailed in~\tabref{data-qh9}.

\paragraph{MD22. }To evaluate the applicability of the Hamiltonian prediction model on large-scale molecules, we adopt the MD22 dataset~\cite{chmiela2023accurate} and focus on the Ac-Ala3-NHMe (ALA3) and DHA molecules. Since the MD22 does not provide Hamiltonian labels (and energy and force labels are provided under a different exchange-correlation functional PBE), we randomly sample 500 structures for each molecule as our benchmark and use PySCF~\cite{sun2018pyscf} to generate Hamiltonian matrices for these structures with the B3LYP exchange-correlation functional and the Def2SVP GTO basis set.

\subsection{DFT Implementation Details}
\label{appx:exp-set-dft}
For this study, all DFT calculations, including those for evaluating the SCF acceleration ratio (\secref{exp-gen}-\ref{sec:exp-large}) and for benchmarking the DFT computation cost (\secref{exp-amor}) are performed using the PySCF software~\cite{sun2018pyscf} with its default parameter settings.

\subsection{Hardware Configurations}
\label{appx:exp-set-hardware}
All neural network models are trained and evaluated on a workstation equipped with a Nvidia A100 GPU with 80 GiB memory and a 24-core AMD EPYC CPU, which is also used for DFT calculations. Note that all the computation times reported in the empirical study are benchmarked on this specific hardware configuration to ensure consistency in comparison. However, it is also recognized that both neural network models and DFT computations have the potential to be parallelized and accelerated using multiple GPUs or CPU cores. Given this capability for parallel processing, establishing a perfectly equitable hardware benchmarking environment for both approaches is challenging. 

\subsection{Training Details}
\label{appx:exp-set-train}

\paragraph{Data-Scarce Scenario.}
We first describe the training details utilized in the empirical study of \secref{exp-gen}. For the self-consistency training setting, we set the total training iterations to 200k for three conformational spaces following \citet{yu2023efficient}. The weighting factor $\lambda_\SC$ is set to 10 across all molecules.
Considering that the efficacy of the supervised learning setting might be limited by scarce labeled data, we allocate a higher number of training iterations (\ie, 500k) to this strategy to ensure the model reaches its optimal performance within the constraints of the available labels. For all experimental conditions and datasets, we maintain a consistent batch size of 5. We utilize a polynomial decay learning rate scheduler to modulate the learning rate (LR) during training, where the polynomial power is set to 5 for self-consistency training and 14 for supervised learning based on empirical trials. Notably, the scheduler increases the learning rate gradually during the first 10k warm-up iterations. The learning rate starts at 0 and peaks at a maximum of 1$\e{-3}$ across all training scenarios.
When addressing the supervised training settings with extended labeled data as mentioned in \secref{exp-amor}, we adopt the same training hyperparameters as those used in the supervised learning setting of \secref{exp-gen}, with the singular adjustment of setting the polynomial power to 5.

\paragraph{Out-of-Distribution Scenario.}
As outlined in \secref{exp-gen}, to benchmark the OOD generalization performance, we initially train the QHNet model on the QH9-small subset, and then fine-tune the model on unlabeled large molecules. 
We continue to use the polynomial learning rate schedule, which includes a warm-up phase. Additional training hyperparameters are detailed in \tabref{exp-train-qh9}.

\begin{table}[ht]
    \caption{Training hyper-parameters in the OOD scenario.}
    \centering
    \begin{tabular}{lccccc}
    \toprule
    Training Phase & Batch size & Maximum LR & Polynomial Power & Iterations & Warm-up Iterations \\
    \midrule
    Pretraining & 32 & 5$\e{-4}$ & 1 & 300k & 1k \\
    Fine-tuning & 5 & 2$\e{-5}$ & 8 & 200k & 3k \\
    \bottomrule
    \end{tabular}
    \label{tab:exp-train-qh9}
\end{table}

\paragraph{Large-Scale Molecular Systems. }As discussed in \secref{exp-large}, we adopt a two-stage training strategy to generalize the Hamiltonian model to large-scale MD22 structures. 
We still employ the polynomial learning rate schedule with the warm-up stage, and detail other training hyper-parameters in \tabref{exp-train-md22}.

\begin{table}[t]
    \caption{Training hyper-parameters in the large-scale molecular systems.}
    \centering
    \begin{tabular}{lccccc}
    \toprule
    Training Phase & Batch size & Maximum LR & Polynomial Power & Iterations & Warm-up Iterations \\
    \midrule
    Pretraining & 32 & 5$\e{-4}$ & 1 & 400k & 5k \\
    Fine-tuning (ALA3) & 2 & 2$\e{-5}$ & 8 & 100k & 10k \\
    Fine-tuning (DHA) & 1 & 2$\e{-5}$ & 8 & 200k & 10k \\
    \bottomrule
    \end{tabular}
    \label{tab:exp-train-md22}
\end{table}

\begin{table}[t]
    \caption{
    Performance comparison between \texttt{self-con}sistency training, and supervised training using \emph{full extended labels}, in the \emph{data-scarce} scenario (in parallel with \tabref{res-cross}), corresponding to the ending points of \figref{amor-md17} (\texttt{extended-label-online} is close to \texttt{extended-label}).
    }
    \centering
    \small
    \setlength\tabcolsep{4pt}
    \begin{tabular}{lccccccccc}
    \toprule
    Molecule & Setting & $\bfH\,[\mu E_\mathrm{h}]\downarrow$ & $\mathbf{\epsilon}\,[\mu E_\mathrm{h}]\downarrow$ & $\bfC\,[\%]\uparrow$ & $\epsilon_{\mathrm{HOMO}}$$\,[\mu E_\mathrm{h}]\downarrow$ & $\epsilon_{\mathrm{LUMO}}$$\,[\mu E_\mathrm{h}]\downarrow$ & $\epsilon_\Delta$$\,[\mu E_\mathrm{h}]\downarrow$ & SCF Accel.$\,[\%]\downarrow$ \\
    \midrule
    \multirow{2}{*}{Ethanol}  & extended-label & \textbf{58.28} & 986.84 & 99.94 & \textbf{230.20} & 2902.14 & 2723.64 & 63.5 \\
    & label\,+\,\scloss & 95.90 & \textbf{340.56} & \textbf{99.94} & 403.60 & \textbf{1426.20} & \textbf{1370.35} & \textbf{61.5} \\
    \midrule
    Malondi- & extended-label & \textbf{71.45} & 1014.12 & 99.63 & \textbf{199.48} & 414.58 & 415.91 & 66.6 \\
    aldehyde  & label\,+\,\scloss & 86.60 & \textbf{280.39} & \textbf{99.67} & 274.45 & \textbf{279.14} & \textbf{324.37} & \textbf{62.1} \\
    \midrule
    \multirow{2}{*}{Uracil} & extended-label & \textbf{52.53} & \textbf{288.29} & 99.38 & \textbf{306.05} & \textbf{294.54} & 398.08 & 58.1 \\
    & label\,+\,\scloss & 63.82 & 315.40 & \textbf{99.58} & 359.98 & 369.67 & \textbf{388.30} & \textbf{54.5} \\
    \bottomrule
    \end{tabular}
    \label{tab:res-cross-md17}
\end{table}

\section{Additional Experimental Results}
\label{appx:exp-res}

\subsection{Generalization Results}
\label{appx:exp-res-gen}
As noted in \secref{exp-gen}, while supervised training can outperform self-consistency training with adequate computational resources, the advantage in terms of Hamiltonian MAE does not consistently extend to molecular properties. This discrepancy has been observed in the OOD generalization scenario in~\tabref{res-cross} of \secref{exp-gen}. Correspondingly, the results presented in~\tabref{res-cross-md17} show that there exists a comparable trend in the data-scarce scenario.

\subsection{Results of Alternative Model Architectures}
\label{appx:exp-alt-archi}

For further validate the advantage of self-consistency training with alternative architectures, we investigate its benefit using the PhiSNet~\citep{unke2021se} architecture, which is another performant model for Hamiltonian prediction on molecules. 
As shown in \tabref{phis-md17}, the results exhibit the same conclusion as shown in \tabref{res-md17}: compared to the results of supervised training (\texttt{label}), applying self-consistency loss (\texttt{label+self-con}) on unlabeled structures leads to a remarkable improvement across all evaluation metrics. Notably, the MAEs for HOMO $\epsilon_{\rm HOMO} $, LUMO $\epsilon_{\rm LUMO} $ and HOMO-LUMO gap $\epsilon_\Delta$ are reduced by several folds. These results demonstrate the generality of self-consistency training for improving the performance of general architectures.

\begin{table*}[ht]
    \vspace{-0.1in}
    \caption{
    Generalization improvement by self-consistency training on unlabeled data \emph{on various model architectures} in the \emph{data-scarce} scenario (\emph{MD17-Ethanol} Hamiltonian).
    Evaluated on the test split of conformations of the molecule. The setting is in parallel with \tabref{res-md17}.
    }
    \centering
    \small
    \setlength\tabcolsep{4pt}
    \begin{tabular}{lccccccccc}
    \toprule
    Architecture & Setting & $\bfH\,[\mu E_\mathrm{h}]\downarrow$ & $\bfeps\,[\mu E_\mathrm{h}]\downarrow$ & $\bfC\,[\%]\uparrow$ & $\epsilon_{\mathrm{HOMO}}$$\,[\mu E_\mathrm{h}]\downarrow$ & $\epsilon_{\mathrm{LUMO}}$$\,[\mu E_\mathrm{h}]\downarrow$ & $\epsilon_\Delta$$\,[\mu E_\mathrm{h}]\downarrow$ & SCF Accel.$\,[\%]\downarrow$ \\
    \midrule
    \multirow{2}{*}{QHNet}  & label & 160.36  & 712.54  & 99.44 & 911.64 & 6800.84 & 6643.11 & 68.3 \\
            & label\,+\,\scloss & \textbf{75.65} & \textbf{285.49} & \textbf{99.94} & \textbf{336.97} & \textbf{1203.60} & \textbf{1224.86} & \textbf{61.5} \\
    \midrule
    \multirow{2}{*}{PhiSNet}  & label & 116.72 & 2702.13 & 98.63 & 1887.50 & 7954.97 & 6834.62 & 65.69 \\
            & label\,+\,\scloss & \textbf{93.77} & \textbf{475.29} & \textbf{99.91} & \textbf{602.96} & \textbf{1645.25} & \textbf{1689.17} & \textbf{62.87} \\ %
    \bottomrule
    \end{tabular}
    \label{tab:phis-md17}
    \vspace{-0.1in}
\end{table*}

\subsection{The Impact of SCF Iteration Strategies} 
\label{appx:exp-res-init}

As mentioned in \secref{exp}, to illustrate the accuracy of Hamiltonian prediction, we assess its capability for accelerating DFT when using the prediction as initialization. Following previous studies~\cite{yu2023efficient,yu2023qh9}, all DFT calculations are carried out using the PySCF~\citep{sun2018pyscf} software, with the PBE XC functional and the Def2SVP basis set being adopted. The direct inversion in iterative subspace (DIIS)~\citep{pulay1980convergence} iteration strategy is employed, following the defaults.  
The results in Tables~\ref{tab:res-md17}-\ref{tab:res-cross} and~\ref{tab:res-md22} show that the Hamiltonian prediction model leads to a substantial SCF acceleration across various molecular systems, and applying self-consistency training can further improve the acceleration gains. Nevertheless, we find that the iteration strategy can considerably influence SCF convergence, which may diminish the benefit of a more accurate initialization. For example, it is known that DIIS may show a non-monotone iteration behavior, meaning that the Hamiltonian in the next iteration may not be closer to the final solution than the Hamiltonian in the current iteration~\citep{sun2016co,sun2017general}. Hence, even when the initial Hamiltonian (predicted by the model) is closer to the final solution, the Hamiltonian in the next iteration may still be farther away from the solution (``DIIS algorithm does not honor the initial guess well. The optimization procedure may lead the wavefunction anywhere in the variational space''~\citep{sun2016co}). To verify this point, we attempt to use the second-order SCF (SOSCF) iteration strategy~\citep{sun2017general} in place of DIIS for running the SCF iteration and summarize the results in \tabref{rebut-qh9}. SOSCF directly engages in orbital optimization, hence guaranteeing monotonicity. In the evaluation setting of OOD generalization on QH9-large test molecules (in parallel with \tabref{res-cross}), we observe a 57.8\% and 56.2\% SCF acceleration for the \texttt{extended-label} and \texttt{self-con} settings respectively. The speedup is indeed improved to the DIIS speedup of 65.0\% and 64.5\% for the respective settings, justifying the speculation. Moreover, in the evaluation setting of large-scale generalization on ALA3 and DHA structures from MD22 (in parallel with \tabref{res-md22}), self-consistency training achieves a 47.5\% and 37.0\% SCF acceleration respectively, substantially better than DIIS speedup of 64.7\% and 67.0\%. These results support that employing the SOSCF convergence method can better honor the quality of the initial guess. Additionally, for DHA structures where the low-quality \texttt{zero-shot} prediction results in a deceleration (170.8\%), SOSCF can lead to worse convergence performance (231.8\%). Notably, the observed speedup appears to be more pronounced on the larger molecular systems (\eg, ALA3 and DHA), which indicates that molecular systems listed in~\tabref{res-cross} may be already easy to converge with DFT and thereby difficult to accelerate further.

\begin{table*}[t]
    \caption{SCF acceleration performance under two SCF iteration strategies, DIIS and SOSCF. (Results in the main paper are under DIIS.) Evaluated on the QH9-large test split in the \emph{OOD} scenario (supervised training uses full extended labels; same setting as \tabref{res-cross}) and on the MD22 molecules in the larger-scale generalization scenario (same setting as \tabref{res-md22}).
    }
    \centering
    \small
    \setlength\tabcolsep{4pt}
    \begin{tabular}{lccc}
    \toprule
    Evaluation setting & Model setting & Iteration Strategy & SCF Accel.$\,[\%]\downarrow$ \\
    \midrule
     \multirow{4}{*}{QH9-large} & \multirow{2}{*}{extended-label} & DIIS & 65.0 \\
      & & SOSCF & 57.8 \\
      \cmidrule{2-4}
     & \multirow{2}{*}{self-con} & DIIS & 64.5 \\
     & & SOSCF & \textbf{56.2} \\
     \midrule
     \multirow{4}{*}{MD22 (ALA3)} & \multirow{2}{*}{zero-shot} & DIIS & 84.6 \\
     & & SOSCF & 71.8 \\
     \cmidrule{2-4}
     & \multirow{2}{*}{self-con} & DIIS & 64.7 \\
     & & SOSCF & \textbf{47.5} \\
     \midrule
     \multirow{4}{*}{MD22 (DHA)} & \multirow{2}{*}{zero-shot} & DIIS & 170.8 \\
     & & SOSCF & 231.8 \\
     \cmidrule{2-4}
     & \multirow{2}{*}{self-con} & DIIS & 67.0 \\
     & & SOSCF & \textbf{37.0} \\
    \bottomrule
    \end{tabular}
    \label{tab:rebut-qh9}
    \vspace{-0.1in}
\end{table*}

\subsection{Amortization Effect of Self-Consistency Training}
\label{appx:exp-res-amor}
As mentioned in \secref{exp-amor}, we directly access the amortization effect by comparing the computational cost of self-consistency training with that of DFT for solving a bunch of structures. It should be noted that measuring the computational cost of DFT on all unlabeled training structures is impractical, thus we run DFT on 50 randomly picked structures for each molecule. The mean computation time derived from these 50 structures serves as a benchmark to approximate the overall computational time required for the complete set of structures.

\begin{figure}[ht]
    \centering
    \includegraphics[width=0.97\linewidth]{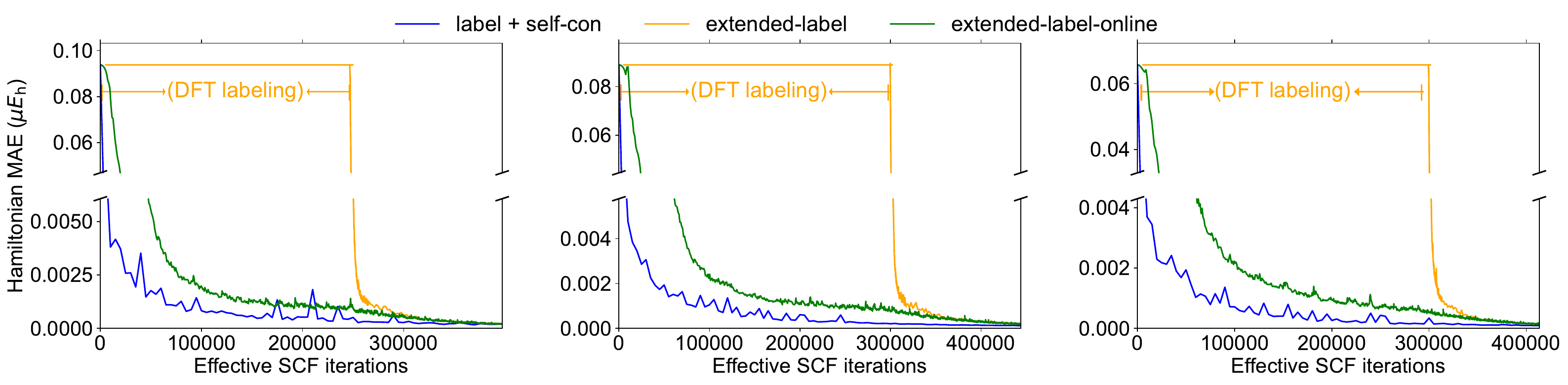}
    \caption{
    Efficiency comparison in the \emph{data-scarce} scenario (MD17 Hamiltonian) among \texttt{self-con}sistency training on unlabeled data, supervised training following DFT labeling on unlabeled data (\texttt{extended-label}), and supervised training along with DFT labeling (\texttt{extended-label-online}). The setting is in parallel with \figref{amor-md17}, with the only difference that
    the cost is measured by \emph{the effective number of SCF iterations} consumed along the training process.
    }
    \label{fig:amor-md17-effscf}
\end{figure}

\begin{figure}[ht]
    \centering
    \includegraphics[width=0.45\linewidth]{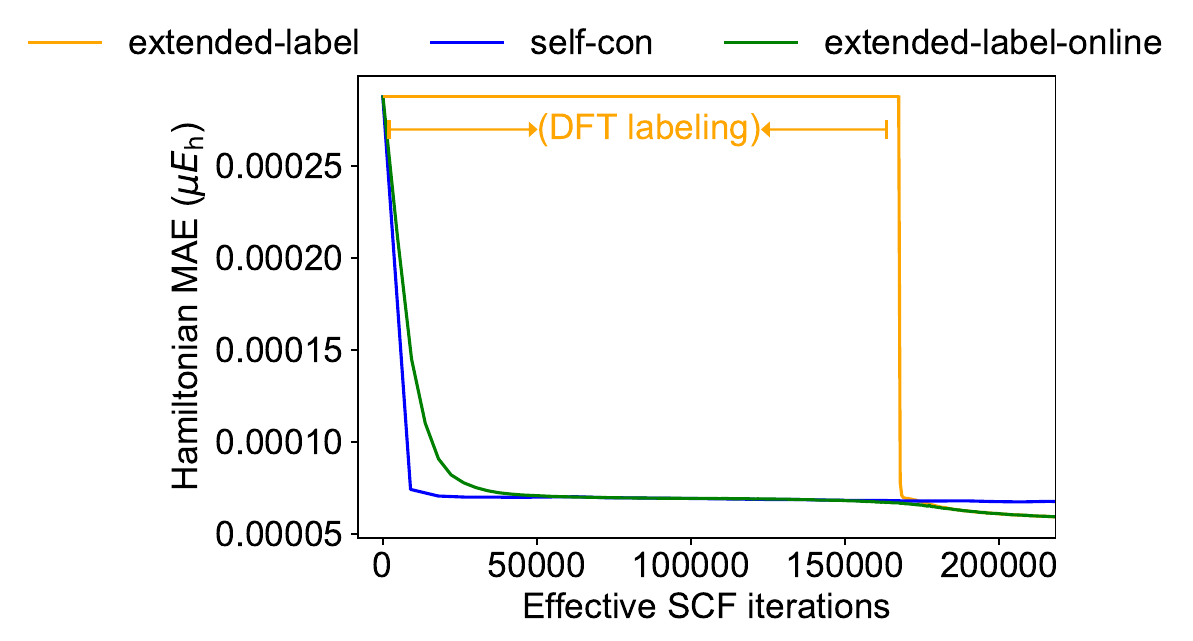}
    \caption{
    Efficiency comparison in the \emph{OOD} scenario (QH9) among fine-tuning using \texttt{self-con}sistency training on unlabeled data, supervised training following DFT labeling on unlabeled data (\texttt{extended-label}), and supervised training along with DFT labeling (\texttt{extended-label-online}).
    The setting is in parallel with \figref{amor-qh9}(b) using the \texttt{adapter} fine-tuning strategy, with the only difference that the cost is measured by \emph{the effective number of SCF iterations} consumed along the training process.
    }
    \label{fig:amor-qh9-effscf}
\end{figure}

To further demonstrate the amortization efficiency of self-consistency training, we also measure the computational cost by ``the number of consumed SCF iterations'' and present the accuracy-cost curves in Figs.~\ref{fig:amor-md17-effscf} and~\ref{fig:amor-qh9-effscf}. The results indicate the same conclusion as shown in Figs.~\ref{fig:amor-md17} and~\ref{fig:amor-qh9}: self-consistency training can achieve a satisfying prediction accuracy even with less SCF steps than the number of SCF steps in the DFT calculation for labeling the molecular structures. Even in the `extended-label-online' setting where the data is generated along with the training of the model, self-consistency training can still achieve a better accuracy given the same budget of SCF iterations.

\subsection{More Results of SCF Acceleration}
\label{appx:exp-res-accel}

\begin{figure}[t]
    \centering
    \subfigure[Comparison of SCF acceleration on \emph{MD17-ethanol} structures]{
        \begin{minipage}[t]{0.6\textwidth}
         \centering
         \includegraphics[width=\textwidth]{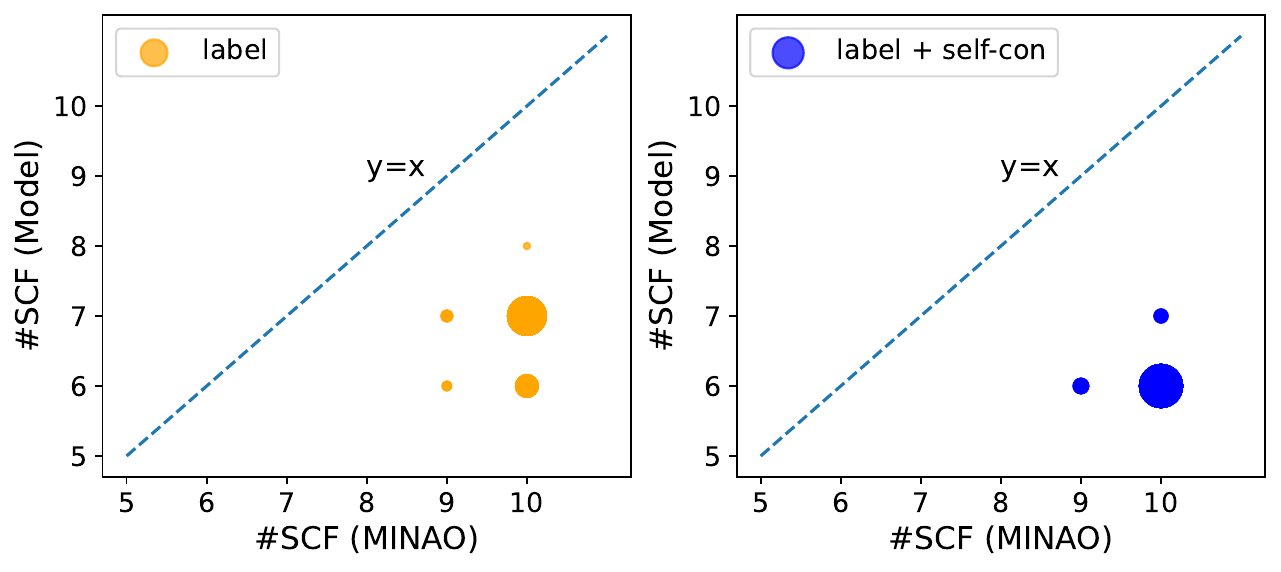}
         \label{fig:p2p-eth}
     \end{minipage}
    }
    \subfigure[Comparison of SCF acceleration on \emph{MD17-malondialdehyde} structures]{
        \begin{minipage}[t]{0.6\textwidth}
         \centering
         \includegraphics[width=\textwidth]{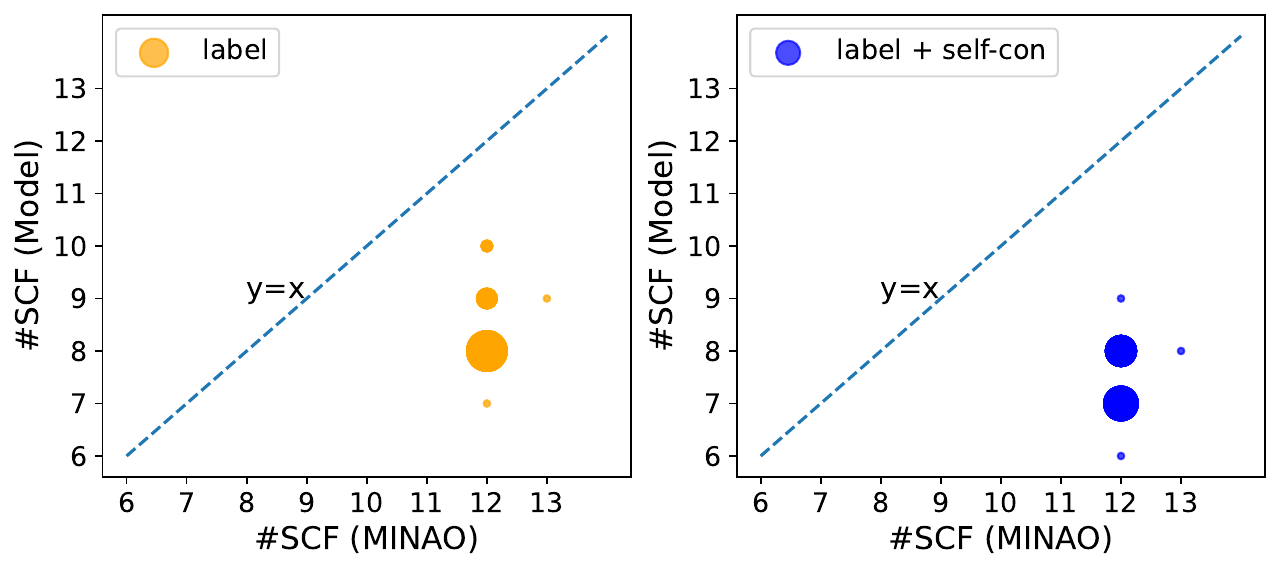}
         \label{fig:p2p-mal}
     \end{minipage}   
    }
    \subfigure[Comparison of SCF acceleration on \emph{MD17-uracil} structures]{
        \begin{minipage}[t]{0.6\textwidth}
         \centering
         \includegraphics[width=\textwidth]{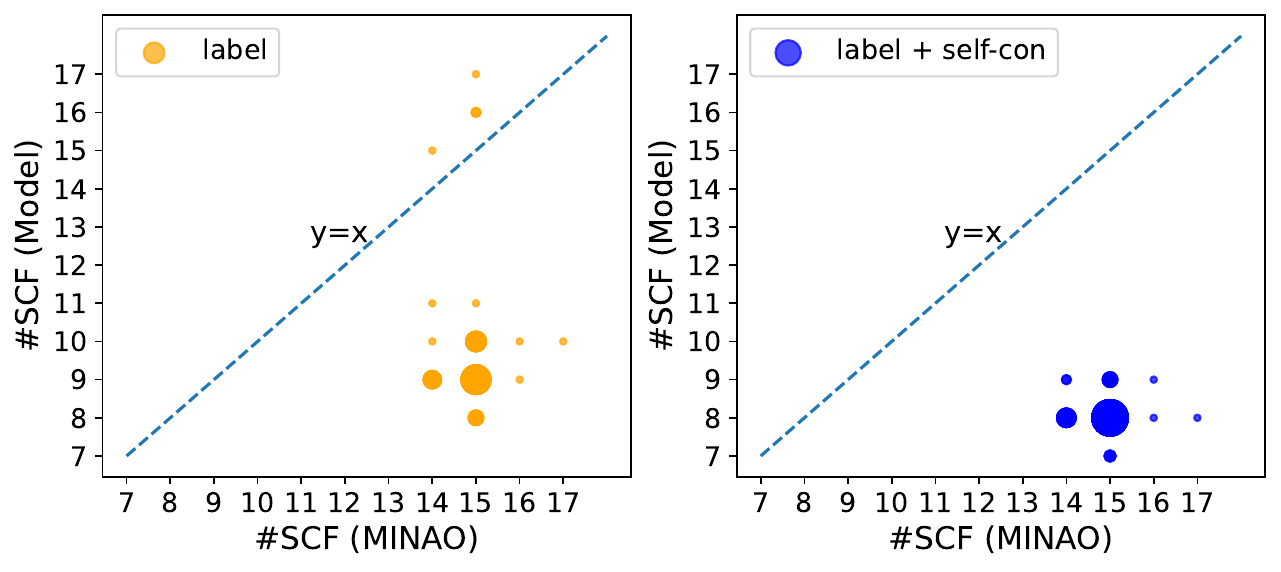}
         \label{fig:p2p-ura}
     \end{minipage}   
    }
    \caption{Comparison of SCF acceleration on \emph{MD17} structures (in parallel with \tabref{res-md17}). Each sub-figure shows a scatter plot of the number of converged SCF steps from two initial guesses: MINAO (\textbf{x}-axis), and the predicted Hamiltonian (\textbf{y}-axis) by a model trained using labels (left) and additionally using self-consistency training (right). All figures are plotted using 50 data points.}
    \label{fig:p2p-md17}
\end{figure}

\begin{figure}
    \centering
    \includegraphics[width=0.9\linewidth]{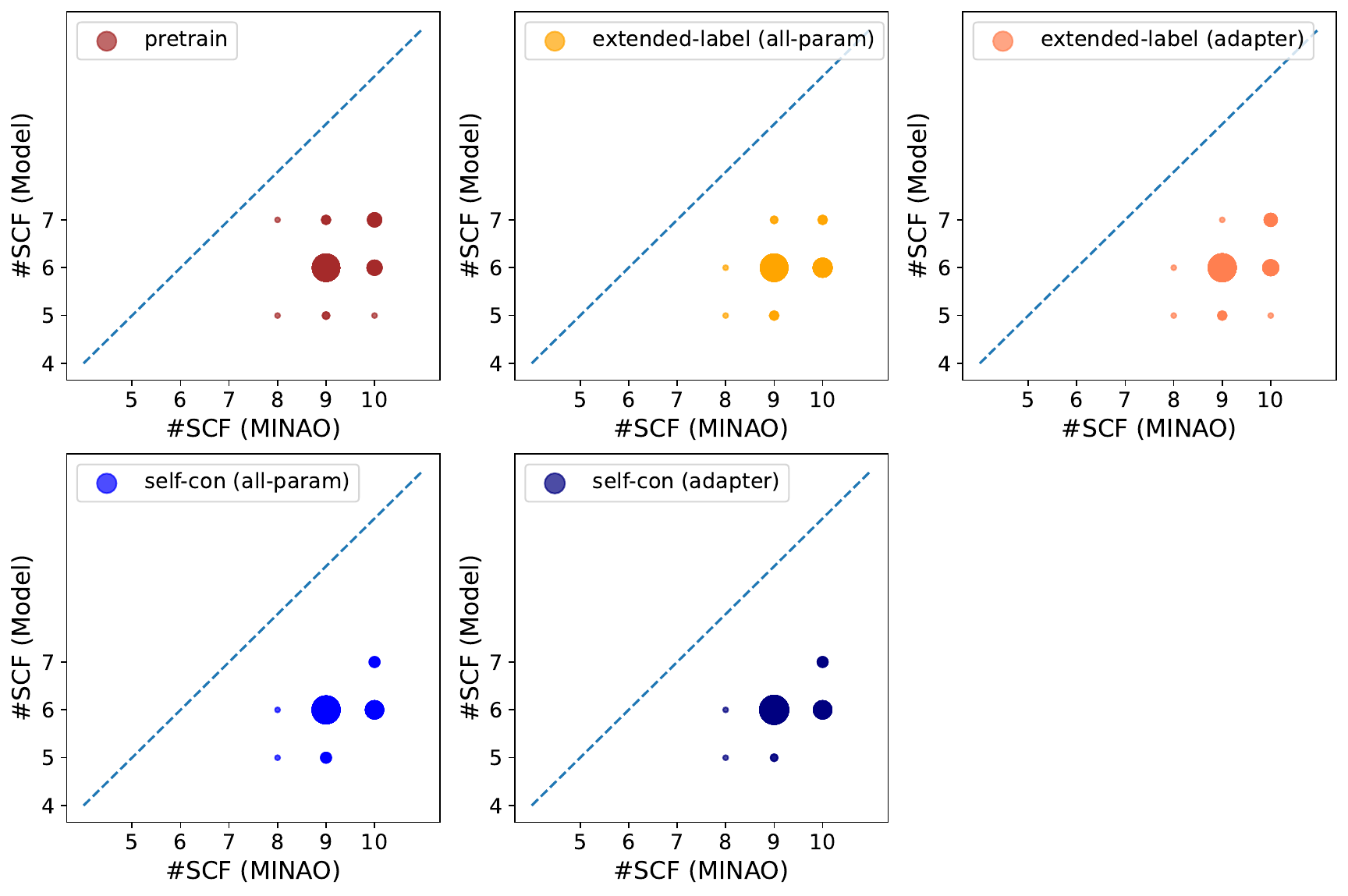}
    \caption{Comparison of SCF acceleration on \emph{QH9-large} structures (in parallel with \tabref{res-cross}). Each figure shows a scatter plot of the number of converged SCF steps from two initial guesses: MINAO (\textbf{x}-axis), and the predicted Hamiltonian (\textbf{y}-axis) by a model pretrained using labels (top left) and additionally finetuned using labels (top middle and right) or self-consistency loss (bottom left and middle) with two fine-tuning strategies. All figures are plotted using 50 data points.}
    \label{fig:p2p-qh9}
\end{figure}

\begin{figure}[t]
    \centering
    \subfigure[Comparison of SCF acceleration on \emph{MD22-ALA3} structures]{
        \begin{minipage}[t]{0.6\textwidth}
         \centering
         \includegraphics[width=\textwidth]{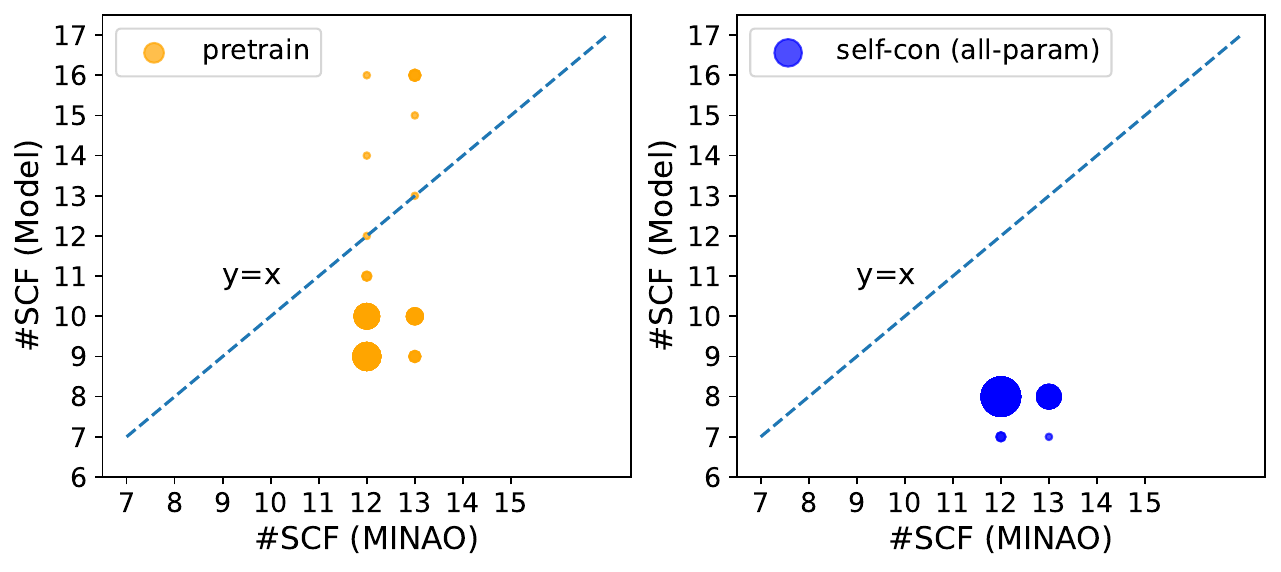}
         \label{fig:p2p-ala3}
     \end{minipage}
    }
    \subfigure[Comparison of SCF acceleration on \emph{MD22-DHA} structures]{
        \begin{minipage}[t]{0.6\textwidth}
         \centering
         \includegraphics[width=\textwidth]{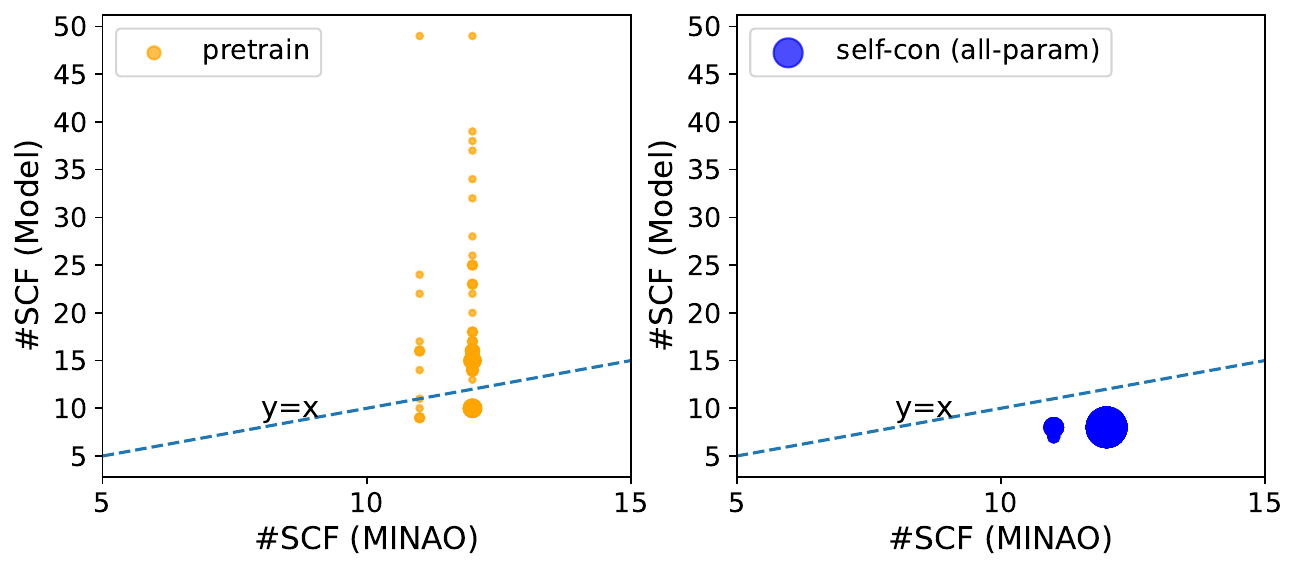}
         \label{fig:p2p-dha}
     \end{minipage}   
    }
    \caption{Comparison of SCF acceleration on \emph{MD22} structures (in parallel with \tabref{res-md22}). Each sub-figure shows a scatter plot of the number of converged SCF steps from two initial guesses: MINAO (\textbf{x}-axis), and the predicted Hamiltonian (\textbf{y}-axis) by a model pretrained using labels (left) and additionally finetuned using self-consistency loss with the \texttt{all-param} strategy (right). All figures are plotted using 50 data points.}
    \label{fig:p2p-md22}
\end{figure}

To comprehensively investigate the significance of our method in accelerating SCF convergence, we present a detailed point-by-point comparison of SCF acceleration for different Hamiltonian prediction models. The results on three datasets are summarized in Figs.~\ref{fig:p2p-md17}-\ref{fig:p2p-md22}. Remarkably, the models with self-consistency training always lead to faster convergence than conventional MINAO guess across various settings. In contrast, the label-based training method for uracil in~\tabref{res-md17} (see Fig.~\ref{fig:p2p-md17}(c)) and the \texttt{zero-shot} setting for two MD22 molecules in~\tabref{res-md22} (see \figref{p2p-md22}) result in slower convergence than MINAO on some structures, while applying self-consistency training can eliminate this issue.

\subsection{Large-Scale Generalization Results}
\label{appx:exp-res-large}

As discussed in \secref{exp-large}, we assess the generalization of self-consistency training to large-scale MD22 structures by comparing it with two state-of-the-art end-to-end (\texttt{e2e}) property prediction models. For this purpose, we choose two advanced \texttt{e2e} architectures, ET\footnote{\href{https://github.com/torchmd/torchmd-net}{https://github.com/torchmd/torchmd-net}, the code is freely available under the terms of the MIT license.}~\cite{tholke2021equivariant} and Equiformer\footnote{\href{https://github.com/atomicarchitects/equiformer}{https://github.com/atomicarchitects/equiformer}, the code is freely available under the terms of the MIT license.}~\cite{liao2022equiformer}, as our baselines.
They are representatives for equivariant architectures that utilize vector features and high-order tensor features, respectively. 
Despite the availability of pretrained models for these methods, they are not directly applicable for our analysis because they were originally trained on the QM9 dataset, which differs from the QH9 dataset used in our study. The two datasets use distinct orbital basis sets for DFT calculations, resulting in slightly different label distributions. 
To ensure a fair comparison, we retrain the ET and Equiformer models on the QH9-full training split, with the default hyper-parameter settings specified in their codebases.
The performance of the two \texttt{e2e} predictors on the QH9-full validation set, as shown in~\tabref{appx-res-md22}, aligns with the outcomes reported in their respective original publications, validating the effectiveness of our model replication. The substantial performance disparity between the QH9-full validation set and two MD22 molecules implies a significant generalization challenge for \texttt{e2e} predictors. Fortunately, this challenge can be potentially alleviated by applying self-consistency training to Hamiltonian prediction.

\begin{table}[t]
    \caption{
    Generalization results of \texttt{e2e} property predictors on \emph{larger-scale} MD22 molecules.
    Models are pretrained on the QH9-full training split and directly evaluated on the large-scale molecules. The \texttt{e2e} predictors significantly suffer from the generalization gap. QH9(valid) denotes the QH9-full validation split. 
    }
    \centering
    \small
    \setlength\tabcolsep{4pt}
    \begin{tabular}{lccccccccc}
    \toprule
     Model & \multicolumn{3}{c}{$\epsilon_{\mathrm{HOMO}}$$\,[\mu E_\mathrm{h}]\downarrow$} & \multicolumn{3}{c}{$\epsilon_{\mathrm{LUMO}}$$\,[\mu E_\mathrm{h}]\downarrow$} & \multicolumn{3}{c}{$\epsilon_\Delta$$\,[\mu E_\mathrm{h}]\downarrow$} \\
    \cmidrule(lr){2-4} \cmidrule(lr){5-7} \cmidrule(lr){8-10}
    & QH9 (valid) & ALA3 & DHA & QH9 (valid) & ALA3 & DHA & QH9 (valid) & ALA3 & DHA \\
    \midrule
    e2e (ET) & 818.77 & 1.74$\times10^{5}$ & 2.92$\times10^{5}$ & 540.22 & 7.72$\times10^{4}$ & 2.58$\times10^{4}$& 1.38$\e{3}$ & 2.38$\times10^{5}$ & 3.39$\times10^{5}$ \\
    e2e (Equiformer) & 646.42 & 2.38$\times10^{5}$ & 3.76$\times10^{5}$& 488.40 & 1.16$\times10^{4}$ & 2.31$\times10^{4}$ & 1.15$\e{3}$ &2.27$\times10^{6}$ & 4.17$\times10^{6}$ \\
    \bottomrule
    \end{tabular}
    \label{tab:appx-res-md22}
    \vspace{-0.1in}
\end{table}

\section{Limitations and Future work} 
\label{appx:limit}

Even though self-consistency training has shown improved efficiency than conventional DFT calculation for training a Hamiltonian prediction model or for solving a bunch of molecular structures (\secref{exp-amor}) by amortizing the cost of SCF iterations over queried molecular structures, the computational complexity remains the same as that of DFT.
This complexity may still limit the applicability (but to a higher level) of Hamiltonian prediction to large molecular systems such as biomacromolecules. It would be a promising future work to reduce the complexity of evaluating the self-consistency loss by leveraging techniques from linear-scaling DFT algorithms.
The Hamiltonian prediction model we used in this study, although is already more efficient than a few alternatives, still %
requires considerable cost to evaluate, due to \eg the use of computationally expensive tensor product operations. This calls for designing more efficient neural network architectures for Hamiltonian prediction.
Moreover, current Hamiltonian prediction models only support prediction under a specific basis set, which has a restricted flexibility to trade-off efficiency and accuracy, and is hard to leverage data under different choices of basis. A possible solution to this restriction is including the overlap matrix into the model input, which conveys information about the basis set in a form relevant to the given molecular structure.
%


\end{document}